\documentclass[journal]{IEEEtran}
\usepackage{float}
\usepackage{graphicx}
\usepackage{bm} 

\usepackage[ruled,norelsize]{algorithm2e}
\usepackage{algpseudocode}
\usepackage{amsmath}
\usepackage{xcolor}
\usepackage{color}

\usepackage{booktabs}
\usepackage{threeparttable}
\usepackage{multirow}
\usepackage{amssymb}

\usepackage{ragged2e}

\makeatletter
\newcommand{\removelatexerror}{\let\@latex@error\@gobble}
\makeatother

\usepackage[colorlinks, citecolor=green]{hyperref}

\ifCLASSOPTIONcompsoc
\usepackage[nocompress]{cite}
\else
\usepackage{cite}
\fi

\ifCLASSOPTIONcompsoc
\usepackage[caption=false,font=normalsize,labelfon
t=sf,textfont=sf]{subfig}
\else
\usepackage[caption=false,font=footnotesize]{subfig}
\fi

\ifCLASSINFOpdf
\else
\fi
\begin{document}
	\title{A Federated Data-Driven Evolutionary Algorithm}
	%
	%
	%
	
	\author{
		\thanks{Manuscript received xx, 2021; revised xxx, 2021. The work is supported by the National Natural Science Foundation of China (Basic Science Center Program: 61988101), International (Regional) Cooperation and Exchange Project (61720106008) and National Natural Science Fund for Distinguished Young Scholars (61725301). (\textit{Jinjin Xu and Yaochu Jin contributed equally to this work.}) (\textit{Corresponding authors: Yaochu Jin; Wenli Du.})}
		Jinjin~Xu,
		Yaochu~Jin,~\IEEEmembership{Fellow,~IEEE},
		Wenli~Du,
		Sai Gu
		\IEEEcompsocitemizethanks{
			\IEEEcompsocthanksitem Jinjin Xu, Wenli Du are with the Key Laboratory of Advanced Control and Optimization for Chemical Processes, Ministry of Education, East China University of Science and Technology, Shanghai, 200237, China. E-mail: jin.xu@mail.ecust.edu.cn, wldu@ecust.edu.cn.
			\IEEEcompsocthanksitem Yaochu Jin is with the Department of Computer Science, University of Surrey, Guildford, GU2 7XH, UK. E-mail: yaochu.jin@surrey.ac.uk.
			\IEEEcompsocthanksitem Sai Gu is with the Department of Chemical and Process Engineering, University of Surrey, Guildford, GU27XH, UK. E-mail:sai.gu@surrey.ac.uk.
		}
	}
	
	%
	%

	\markboth{Draft}%
	{Shell \MakeLowercase{\textit{et al.}}: Bare Demo of IEEEtran.cls for IEEE Journals}
	%



	\maketitle
	
	\begin{abstract}
		Data-driven evolutionary optimization has witnessed great success in solving complex real-world optimization problems. However, existing data-driven optimization algorithms require that all data are centrally stored, which is not always practical and may be vulnerable to privacy leakage and security threats if the data must be collected from different devices. To address the above issue, this paper proposes a federated data-driven evolutionary optimization framework that is able to perform data driven optimization when the data is distributed on multiple devices. On the basis of federated learning, a sorted model aggregation method is developed for aggregating local surrogates based on radial-basis-function networks. In addition, a federated surrogate management strategy is suggested by designing an acquisition function that takes into account the information of both the global and local surrogate models. Empirical studies on a set of widely used benchmark functions in the presence of various data distributions demonstrate the effectiveness of the proposed framework.
		
	\end{abstract}
	

	%
	\IEEEpeerreviewmaketitle

	\section{Introduction}
	\label{sec:introduction}

\IEEEPARstart{E}{volutionary} algorithms (EAs), as meta-heuristic techniques, have been shown to be effective solvers for many real-world problems over the past few decades \cite{fleming2002evolutionary}\cite{tapia2007applications} \cite{dasgupta2013evolutionary}\cite{jin2018data}. Apart from their strong capability of tackling non-convex and multi-model problems, EAs do not reply on analytic and differential objective functions and are able to perform optimization on the basis of collected data, which are usually referred to as data-driven optimization \cite{jin2018data}. In some data-driven optimization problems, data are collected through time-consuming numerical simulations or expensive physical experiments. For example, a single run of fire dynamics simulation may take several hours \cite{mouilleau2009cfd}. In these cases, only a limited amount of data is available for data-driven optimization, posing major challenges to building surrogates needed for data-driven optimization. Therefore, existing work on data-driven surrogate-assisted evolutionary optimization focuses on developing surrogate modeling and management techniques \cite{jin2011surrogate} \cite{chugh2019survey} that can help find acceptable solutions with a limited computational budget for single-objective optimization \cite{jin2002framework} \cite{zhou2007combining} \cite{sun2017surrogate} \cite{wang2017committee}, multi-objective optimization \cite{knowles2006parego}\cite{zhang2010expensive} \cite{Lim2010generalizing}, and many-objective optimization \cite{chugh2016surrogate}\cite{habib2019multiple} \cite{guo2020edn}. 

	
One implicit assumption made in most existing work on data-driven evolutionary optimization is that all data for modeling the surrogates is centrally stored, which does not hold for many real-world optimization problems. For example, there are many large-scale complex systems in manufacturing and process industries \cite{li2021surrogate} \cite{mao2019} consisting of sub-systems that may be distributed on multiple locations and all these systems must be considered at the same time to achieve the optimal performance. Collecting data from sub-systems and storing the data on a central server not only give rise to communication problems, but also raise security and privacy concerns. To address the above problem, some research on distributed optimization has been carried out in the field of automatic control, where distributed or decentralized gradient based optimization methods have been developed \cite{shi2015extra} \cite{lian2017can}\cite{zhang2019asyspa}. These work assumes, however, that exact analytic objective functions are available. Most recently, Li et al. \cite{li2021surrogate} present a gradient-based distributed optimization method for black-box optimization problems, under the assumption that the approximated objective functions are differentiable and strongly convex. Thus, their optimization algorithm is not applicable to optimization problems such as airfoil design \cite{liu2018survey} and trauma system design \cite{wang2016data}. 

In evolutionary computation, distributed evolutionary algorithms (EAs) have been investigated to reduce computation time or to deal with large-scale optimization problems. For example, parallel evolutionary optimization \cite{harada2020survey} based on a master-slave mode \cite{Cantu-Paz97}, island mode \cite{michel1998island} and grid mode \cite{lim2007} have been proposed to perform fitness evaluations in a parallel and distributed manner to reduce the required computation time, assuming that multiple processors are available. In addition, a large number of cooperative and co-evolutionary algorithms \cite{ma2018} have been proposed for solving large-scale and complex optimization problems, which can largely be divided into population-distributed \cite{folino2008training}\cite{roy2009distributed} and dimension-distributed \cite{bouvry2000distributed}\cite{subbu2004network}. To further reduce the computation time, a surrogate model is built for each sub-population in \cite{ren2019surrogate}. A comprehensive survey of distributed EAs, including parallel, hierarchical and co-evolutionary algorithms can be found in \cite{gong2015distributed}. It should be noted that none of these distributed EAs are meant for solving data-driven optimization problems where the data is distributed on multiple devices.  

Meanwhile, a distributed data-driven machine learning paradigm, called federated learning \cite{mcmahan2016communication} \cite{yang2019survey}, has received increasing attention in the field of machine learning. In federated learning, multiple clients collaboratively train a global model without requiring to upload the data collected on multiple clients to a server, reducing the privacy and security risks. Interestingly, evolutionary algorithms (EAs) have been applied in federated learning, including using EAs to optimize the mixture coefficients or model weights \cite{mohri2019agnostic}\cite{li2018federated}, or performing evolutionary neural architecture search in the federated learning framework \cite{zhu2020multi}\cite{zhu2020survey}. To the best of our knowledge, however, none of the above work deals with distributed data-driven surrogate-assisted evolutionary optimization, where surrogates are built on the basis of data distributed collected on multiple devices.     

This work aims to propose a framework for federated data-driven optimization to address a class of distributed data-driven optimization problems, focusing on surrogate construction and surrogate management in a distributed environment in the presence of possibly noisy and non-independently identically distributed (non-iid) data. The main contributions of the work are summarized as follows.
	
\begin{enumerate}
\item On the basis of federated learning, a surrogate-assisted federated data-driven evolutionary algorithm, called FDD-EA, is proposed, which does not require to store data on a single server that are originally collected on multiple devices.
\item A sorted averaging method is designed for aggregating local radial-basis-function networks into a global surrogate on the server, thereby enhancing the performance of incremental federated learning in particular in the presence of non-iid data. 
\item The lower confidence bound acquisition function is adapted to the federated optimization environment, which integrates information from both local and the global surrogate models.
\end{enumerate}

To validate the proposed federated data-driven evolutionary optimization framework, benchmark optimization problems are adopted for generating data distributed on multiple machines. Non-iid data are simulated by assuming that each device is not able to generate data in some decision subspaces. Our experimental results demonstrate that the proposed federated data-driven optimization framework comparably well or better than the state-of-the-art centralized online data-driven surrogate-assisted evolutionary algorithms on the majority of the tested instances.   

The remainder of this paper is organized as follows. In Section \ref{sec:2}, we briefly review the basic federated learning paradigm and a few widely used surrogate-assisted optimization approaches, on both of which the proposed work is based. Section \ref{sec:3} presents the proposed federated data-driven evolutionary optimization framework. In Section \ref{sec:4}, the benchmark problems used in the empirical studies, experimental settings and the comparative results are given. Finally, conclusions are drawn and future directions are discussed in Section \ref{sec:5}.
	
\section{Related Work}
\label{sec:2}
In this section, we briefly review the background of this work, including federated learning, radial basis function network, and the acquisition functions for surrogate-assisted evolutionary algorithms.
	
\subsection{Federated Learning}
The explosion of data resulting from massive edge devices, e.g., Internet-of-Things (IoT), mobile phones and enterprise clouds, has led to the emergence of many edge computing algorithms and frameworks. Federated learning is one of the most prevalent edge computing methods due to its capability of privacy preservation, data security and communication efficiency. The concept of federated learning was first proposed in \cite{konevcny2016federated}\cite{bonawitz2019towards}, and a large body of work has been reported to further reduce communication cost \cite{sattler2019robust} \cite{zhu2020multi} \cite{xu2020ternary}, handle vertical data partition \cite{chen2020vafl}, or enhance privacy protection \cite{triastcyn2019federated}\cite{wei2020federated}\cite{zhu2020encrypt}. However, the test accuracy of federated learning usually suffers from the non-iid nature of the client data compared  to centralized machine learning methods, since the client data may be drawn from different distributions in real-world applications. To alleviate this problem, a large number of new techniques have been proposed, including data sharing \cite{zhao2018federated}, client selection \cite{nishio2019client}, and adaptive federated learning \cite{wang2019adaptive}.
    
\begin{figure}[ht]
		\centering
		\includegraphics[width=3.3in]{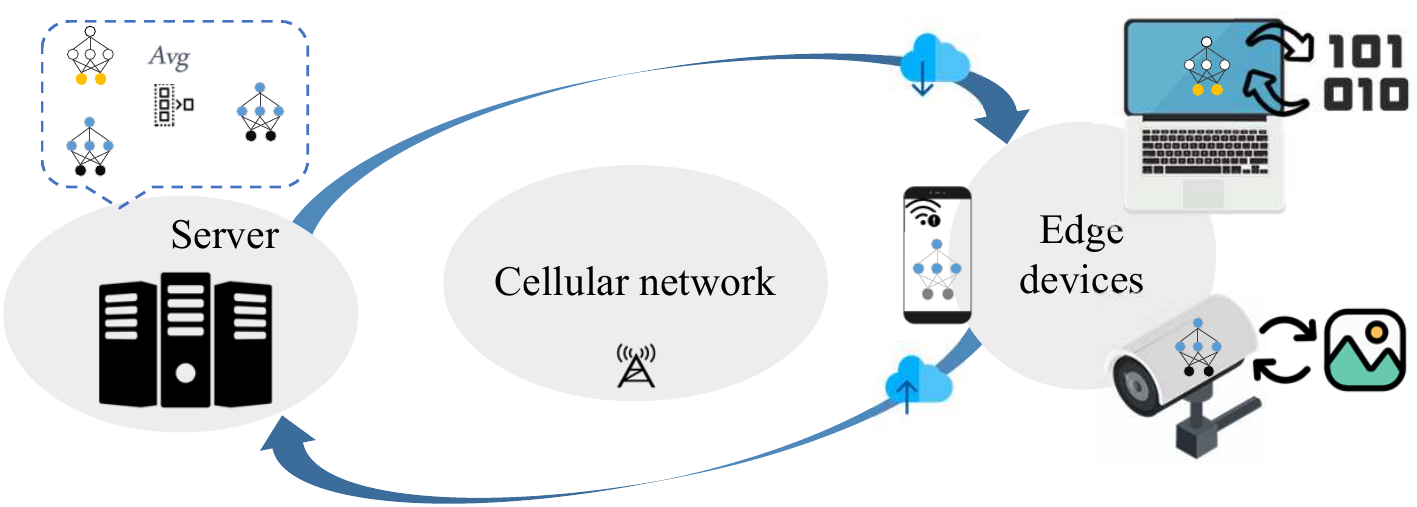}
		\caption{An illustrative example of federated learning. The edge devices train local models on the private local data and send the trained model parameters to the server for aggregation to obtain the global model. Then, all clients download the aggregated global model and repeat the process until the global model converges.}
		\label{fig:federated-learning}
\end{figure}
    
As illustrated in Fig. \ref{fig:federated-learning}, a vanilla (standard) federated learning system consists of a server, a communication network and a number of edge devices, also called local clients \cite{mcmahan2016communication}. In the first round, the server randomly initialize a global model and send the model to all participating local clients. Note that not all clients participate in each round of model update. Then, each participating local client trains the received global model with its own local data, typically using the gradient-based method, for a number of epochs. After training, the participating local clients uploads its updated local model to the server. Finally, the server will aggregate the uploaded local models using weighted averaging, known as FedAvg \cite{mcmahan2016communication}, and send the the updated global model to the participating local clients for the next round of model update. This process repeats until the global model converges. 

Assume in the current round, the global model parameters $\bm{w}$ are sent to $\lambda N$ participating clients, where $\lambda$ is the ratio of the local clients participating in the current round and $N$ is the total number of local clients. Then the $k$-th client trains the received model on the local dataset $D_k$ consisting of $n_k$ training data pairs ($\bm{x}_i$, $y_i$), where $i=1,2,3,...,n_k$, resulting in the updated local model $\bm{w}_k$. The loss function $F_k$ of the local client $k$ is defined by
	\begin{equation}
	\label{eq:loss_fun}
	F_k(\bm{w}_k) = \frac{1}{n_k}\sum_i^{n_k}L(\bm{x}_i,y_i;\bm{w}_k) + \gamma (\bm{w}_k),
	\end{equation}
where $L(\cdot)$ is the user-defined loss function and $\gamma$ denotes the regularizer. And the aim of federated learning is to minimize the global objective $F$ with a global model $\bm{w}$. Therefore, the global loss function of the federated learning system can be defined as:
	\begin{equation}
	\label{eq:fed_loss}
	\min_w \left\{F(\bm{w}) =  \sum_{k=1}^{\lambda N}p_kF(\bm{w}_k) \right\},
	\end{equation}
where $p_k$ is the weight of $\bm{w}_k$ and calculated by
    
\begin{equation}
    \label{eq:p_k}
    p_k = \frac{n_k}{\sum_{k=1}^{\lambda N}n_k}.
\end{equation}
	
In each round, the local model $\bm{w}_k$ is initialized with the downloaded global $\bm{w}$, then client $k$ can update $\bm{w}_k$ using the mini-batch stochastic gradient descent (SGD) \cite{lecun2015deep} with a learning rate $\eta_k$ and performs $E (\ge 1)$ local training epochs:
	
	\begin{equation}
	\label{eq:local_update}
	\bm{w}_k^{i+1} = \bm{w}_k^{i+1} - \eta_k^i \nabla F_k(\bm{w}_k^i), i = 0, 1,..., E-1.
	\end{equation}

At the end of each round, the server aggregates all updated local models to obtain the updated global model and start the next round. 

Note that the vanilla federated learning algorithm adopts a deep neural network and assumes there is enough training data. However, as we have previously discussed, in data-driven evolutionary optimization, the number of training data is usually very limited since data collection is very expensive. Thus, the global model we are going to use will be a tiny model. More discussions will be provided in Section \ref{sec:3}.
	
\begin{figure*}[htb]
		\centering
		\includegraphics[width=6.2in]{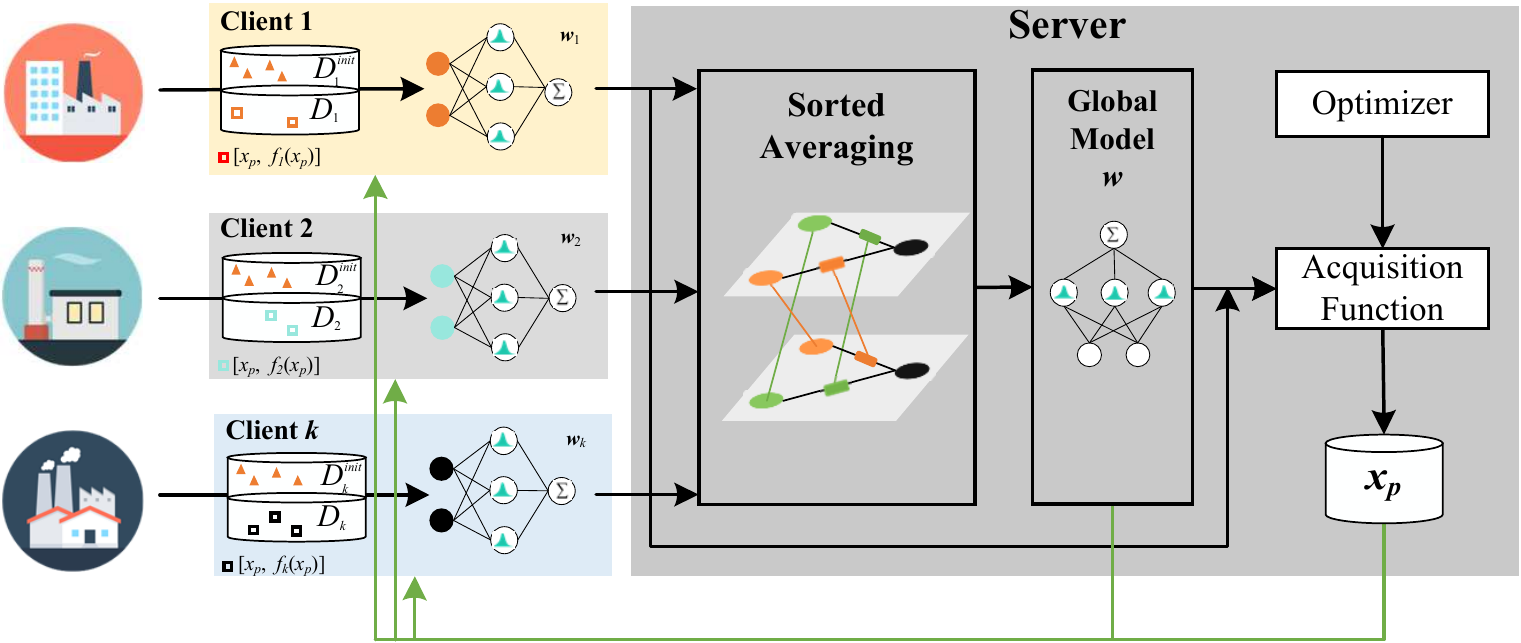}
		\caption{The overall framework of the proposed FDD-EA. Firstly, the clients build local surrogates trained on the private dataset $D_{init}$. Secondly, the server obtains the global surrogate $\bm{w}$ by averaging the received local surrogates using the sorted averaging method, and then conducts evolutionary optimization assisted by the global model $\bm{w}$. Thirdly, promising solutions are selected with the help of a proposed surrogate management strategy, which are broadcast to all clients for data collection, i.e., evaluation of the fitness value using the expensive objective functions $f_k$. Finally, the local surrogates are updated on the union of $D_k^{init}$ and the newly sampled data $D_k$.}
		\label{fig:SA-FedOpt}
\end{figure*}
    
\subsection{Radial-Basis-Function Networks}
Various regression, classification and interpolation methods can be used as the surrogate models to approximate the real objective functions \cite{jin2005comprehensive}. Among them, radial basis function networks (RBFNs) \cite{wang2018offline}, artificial neural networks (ANNs) \cite{jin2002framework}, Gaussian processes (GPs) \cite{liu2013gaussian} and polynomial regression (PR) \cite{wang2017committee} are the most widely ones, and surrogate ensembles have also been widely studied \cite{wang2018offline, goel2007ensemble}. In this work, an RBFN is adopted as the surrogates because it has, based on our pilot studies, shown to be more scalable to the number decision variables and easier to train. The structure of the RBFN is similar to that in \cite{du2014radidal}. Let $\bm{x}_i \in \mathbb{R}^d$ be the $i$-th input sample of the training data, the prediction of the RBFN can be denoted by:
	\begin{equation}
	\label{eq:RBF_pred}
	\hat{y_i} = \sum_{j=1}^m a_j \varphi_j(||\bm{x}_i-\bm{c}_j||) + b,
	\end{equation}
where $m$ is the number of centers of the RBFN, $\bm{a} = (a_1, a_2,\cdots,a_m)$ are the weights of the RBFN (the subscripts here indicate the weight indices of the RBFN), the center vector $\bm{C} = (\bm{c}_1, \bm{c}_2,..., \bm{c}_m)$ is obtained by the k-means clustering algorithm, $\bm{c}_m \in \mathbb{R}^d$, and $b \in R$ is the bias. And $||\bm{x}_i-\bm{c}_j||$ is the Euclidean distance between an input and the center $\bm{c}_j$. Finally, $\varphi_j(\cdot)$ denotes the basis function, and there are various choices for $\varphi_j$, such as the Gaussian function, logistic function, or thin-plate spline function \cite{du2014radidal}. In this work, we use the Gaussian function, which is expressed by
	\begin{equation}
	\label{eq:gaussian}
	\varphi_j(||\bm{x}_i-\bm{c}_j||) = e^{-\frac{||\bm{x}_i-\bm{c}_j||^2}{2\delta_j^2}},
	\end{equation}
where $\delta_j$ denotes the standard deviations, also known as the spreads, width or radii. In this work, we select $\delta_j$ according to the maximum distance between the centers \cite{du2014radidal}.
	
For the weights connecting the hidden nodes and the output of the network, we have
	\begin{equation}
	\label{eq:rbf_y}
	    \resizebox{.9\hsize}{!}{$\begin{bmatrix}
    	    \varphi_1(||\bm{x}_1-\bm{c}_1||)& ...& \varphi_m(||\bm{x}_1-\bm{c}_m||) \\
    	    ... & ... \\
	        \varphi_1(||\bm{x}_{n_k}-\bm{c}_1||)& ...& \varphi_m(||\bm{x}_{n_k}-\bm{c}_m||) \\
    	\end{bmatrix}
    	\begin{bmatrix}
    	    a_1 \\
    	    ... \\
    	    a_m \\
    	\end{bmatrix} +b \approx 
    	\begin{bmatrix}
    	    y_1 \\
    	    ... \\
    	    y_{n_k} \\
    	\end{bmatrix},$}
	\end{equation}
where $n_k$ is the total number of the training samples, and the pseudo-inverse method or gradient-based method \cite{du2014radidal} can be employed to train the weights.
	
\subsection{Acquisition functions}
Surrogate management, which determines which new solutions are to be sampled, i.e., evaluated using the expensive objective functions, is central to the effectiveness of surrogate-assisted evolutionary algorithms (SAEAs) \cite{jin2005comprehensive}. Among different surrogate management strategies, acquisition functions in Bayesian optimization \cite{Bobak2016}, also known as infill criteria in global optimization \cite{kushner1964new,zhilinskas1975single, jones1998efficient}, are mathematically solid and have been shown very effective in balancing exploration and exploitation in online data-driven surrogate-assisted evolutionary optimization. 
    
Several acquisition functions have been proposed to sample new solutions  \cite{auer2002using}, such as lower confidence bound (LCB) \cite{torczon1998using}, expected improvement (EI) \cite{jones1998efficient} and probability of improvement (PI) \cite{vzilinskas1992review}. Given a candidate solution $\bm{x}_p$, LCB of $\bm{x}_p$ is calculated by:
	\begin{equation}
	\label{eq:LCB}
	{\rm LCB}(\bm{x}_p)=\hat f(\bm{x}_p)-\mu \,\hat s(\bm{x}_p),
	\end{equation}
where $\hat f(\bm{x}_p)$ and $\hat s (\bm{x}_p)$ are the predicted mean and standard deviation (the confidence level of the prediction) of the solution point $\bm{x}_p$, respectively. The trade-off hyperparameter $\mu$ is usually set to 2 as recommended in \cite{emmerich2006single}. 

In this work, we adapt the LCB to the federated optimization framework because the original LCB cannot be directly applied. The reason is that the RBFN is adopted as the surrogates and hence the uncertainty information of the predictions, $\hat s(\bm{x}_p)$, is not directly available. Therefore, a federated infill criterion based on LCB is proposed, which will be discussed in greater detail in Section \ref{subsec:F-LCB}.

\section{Problem definition and Proposed Algorithm}
	\label{sec:3}
In this section, we at first formulate the federated data-driven optimization problem to be addressed in this work. Then, the overall workflow of the proposed FDD-EA is described. Finally, a new strategy for aggregating the local RBFN surrogates will be presented together with an adapted LCB.
	
\subsection{Problem definition}
	\label{subsec:problem_def}
	
As stated in Section \ref{sec:introduction}, the main purpose of the present work is to solve data-driven optimization problems where the raw data for optimization is distributed on different local machines and not allowed to be transmitted to a central server. In addition, we assume that each client has all decision variables, and is able to sample a limited amount of new data as informed by the server. However, each client may be limited to sampling solutions in a given subspace of the decision space, resulting in a horizontal but non-iid data distribution on different clients. 
	
Thus, in federated data-driven evolutionary optimization systems considered in this work, only the local clients can sample new data by performing expensive simulations or experiments in a limited sub-space of the whole decision space. Denote the approximated objective function on the $k$-th client by
	\begin{equation}
	\label{eq:expensive_f}
	y_k=f_k(\bm{x}),
	\end{equation}
where $\bm{x} \in \mathbb{R}^d$ is the decision vector. Since we consider horizontal data partition in this work, the expensive objective function of all local clients are the same, although the decision variables of the clients may be limited to a subspace in the overall decision space.
    
On the $k$-th client, a local dataset $D_k = \{(\bm{x}_1,y_1),..., (\bm{x}_{n_k},y_{n_k})\}$ consisting of $n_k$ data pairs of decision vectors and their corresponding fitness values evaluated according to Equation (\ref{eq:expensive_f}) is stored. Thus, a local surrogate model $\bm{w_k}$ can be constructed on $D_k$, which is described by:
    \begin{equation}
	\label{eq:surrogate_f}
	\hat f_k=\bm{w}_k(\bm{x}|D_k).
	\end{equation}
Denote the difference between the surrogate and the real fitness function on the $k$-th client as $\epsilon$, then we have $f_k(\bm{x})=\hat f_k + \epsilon$.
	
Recall that the server has no direct access no the training data describing the functional relationship between a decision vector and its objective function value. Furthermore, the clients are not allowed to upload such raw data to the server, neither can they communicate the local raw data to other clients. However, the clients are allowed to upload the parameters of their local surrogates to the server for constructing a global surrogate. Consequently, the server can aggregate the uploaded local surrogates to build global surrogate, based on which the optimum of the overall system $\bm{x}^*$ can be found using an optimizer, an EA in this work. In other words, the evolutionary search is conducted on the server instead of on the local clients, assuming that the computational power on the clients is limited.
	
Similar to centralized data-driven evolutionary optimization, the global surrogate must be properly managed to effectively assist the EA to find the optimum of the global system. Once a promising solution $\bm{x}_p$ is selected according to the acquisition function on the server, it will be broadcast to all participating clients. If the solution is within the feasible subspace of $k$-th client, the objective value of $\bm{x}_p$ will be sampled using $f_k(\bm{x}_p)$, in practice, a time-consuming numerical simulation will be performed or an experiment will be conducted. The sampled new data pair, $(\bm{x}_p, f_k(\bm{x}_p))$ will be added in the local dataset $D_k$ for updating the local surrogate. If solution $\bm{x}_p$ falls in its infeasible decision subspace, i.e., the $k$-th client is not able to sample this solution, no new data is sampled in this round. Nevertheless, the local surrogate can still be updated in the next round if this client participates the model update.     

The proposed federated data-driven evolutionary optimization problem and federated learning share some common features, but they also have clear differences. The main goal of federated learning is to build a high-quality global model for classification or regression, without requiring the data distributed on multiple clients to be uploaded and centrally stored on the server so that data security can be ensured and privacy can be protected. Similarly, federated data-driven optimization also needs to build a global surrogate model, which is used to assist the evolutionary search of the optimum of the global system. The global surrogate is built in a federated learning manner so that the local data does not need to transmitted to the server to reduce security and privacy issues. However, federated data-driven optimization distinguishes itself with federated learning in at least the following two aspects. First, the goal of federated data-driven optimization is to find the optimum of the whole system, and as a result, a proper surrogate model management strategy must be designed, determining which new solutions should be sampled on the local clients. Similar to centralized surrogate-assisted evolutionary optimization \cite{huesken2005PI}, the capability of effectively guiding the evolutionary search, rather than the prediction accuracy, is of paramount importance in federated data-driven evolutionary optimization. The second main difference is that in federated data-driven optimization, new data keeps being generated and therefore the surrogates must be incrementally updated during the optimization. In contrast to federated learning where the amount of data on each client is often big, the amount of data on the local clients is usually very limited in federated optimization. The main differences between federated data-driven optimization and federated learning are summarized in Table \ref{table:difference}.  

It should also be pointed out that federated data-driven optimization differs from existing parallel or distributed evolutionary optimization in that in the former, the raw data is collected and stored in a distributed way, while in the latter, the computation or optimization is proactively distributed to different machines for reducing computation time. As a result, the way of distributing the computation or data tasks in distributed optimization is under the full control of the user, while in federated optimization, the data distribution is mainly determined by the nature of the local clients (subsystems).

		\renewcommand{\arraystretch}{1.5}
	\begin{table}[H]
		
		
		\centering
		
		\caption{The main differences between vanilla federated learning system and federated data-driven evolutionary optimization.}
		
		\label{table:difference}

		\begin{tabular}{|c|c|c|}
        \hline
         & \begin{tabular}[c]{@{}c@{}}\textbf{Vanilla} \\\textbf{Federated Learning}\end{tabular} & \begin{tabular}[c]{@{}c@{}}\textbf{Federated Data-Driven} \\ \textbf{Evolutionary Optimization} \end{tabular}   \\ \hline
        
        Server     & Model aggregation  & \begin{tabular}[c]{@{}c@{}}Model aggregation and \\ surrogate management\end{tabular} \\ \hline
        
        Local clients     & Local training      & \begin{tabular}[c]{@{}c@{}}Local training\\ and informed sampling\end{tabular}                \\ \hline
        Data & (Usually) Big, stationary & Small, incremental\\ \hline
        
        Objective        & Prediction accuracy                  & Solution optimality              \\ \hline
        \end{tabular}
        
        \vspace{3pt}
		
	\end{table}
	
\subsection{Overall framework}
\label{sec:framework}
The overall framework of FDD-EA is given in Fig. \ref{fig:SA-FedOpt}. In the beginning, the server samples a certain amount of data using the Latin hypercube sampling (LHS) method \cite{stein1987large}. These solutions are sent to all clients and evaluated on the clients using the local real objective function, which constitutes the initial training set $D_k^{init}$ of each client. Note that if not all clients can sample the whole decision space, $D_k^{init}$ can also vary from client to client. Then each client uses $D_k^{init}$ to train a local RBFN surrogate, and uploads the trained model parameters to the server after the training is completed. A global surrogate is obtained by aggregating the local RBFNs using the sorted averaging method, the details of which will be presented in Section \ref{sec:sorted_ave}. An EA is then employed to search for the optimal solutions of the global surrogate by minimizing the proposed federated LCB (to be described in Section \ref{sec:sorted_ave}) for a certain number generations. Finally, the optimal solution found by the EA, $\bm{x}_p$, is broadcast to all clients participating the next round of surrogate update. The point $\bm{x}_p$ will be evaluated on the participating clients using their real objective function $f_k$ and if successful (in their feasible subspace in the non-iid case), be added to their database $D_k$. The surrogate ($\bm{w}_k$) on each participating client will then be updated on the augmented $D_k$. This process repeats until the maximum computation budget is exhausted. The pseudo code of the proposed FDD-EA is given in Algorithm \ref{algo:work}.
	\begin{figure}[htb!]
		\removelatexerror
		\begin{algorithm}[H]
		\caption{Pseudo Code of FDD-EA}
		\label{algo:work}
		\SetKwInOut{Input}{Input}
		\SetKwInOut{Output}{Output}
		\KwIn{Number of participating clients $N$, global surrogate $\bm{w}$, local surrogate $\bm{w}_k$, empty index set $\bm{S}$, local archives $D_{k}$, client weight $p_k$, $k=\{1,2,.., N\}$, maximum number of real objective function evaluations FE$_{max}$. }
		\SetKwProg{Server}{Server does:}{}{end}
		\vspace{3pt}
		\text{{\bfseries Init:} Sample 5$d$ points $\bm{x}_1$, $\bm{x}_2,...,\bm{x}_{5d}$ in the decision}
		\text{space by LHS and evaluate the values $y_1, y_2,...,y_{5d}$}
		\text{with the real objective functions as $D_k^{init}$, and train}
		\text{the initial local models $\bm{w}_k$ with $D_k^{init}$.}
		\vspace{4pt}
		
		\While {\rm{FE} $\le$ FE$_{max}$}
		{
		    \vspace{3pt}
			update $\bm{S} \leftarrow$ $\lambda N$ randomly selected from $N$ clients
			
		    \vspace{3pt}
		    
		    \Server{}
			{   
			    \vspace{3pt}
				/*update the global surrogate*/
				
				\vspace{3pt}
				$\bm{w}$ $\leftarrow$ Algorithm \ref{algo:sorted_average}
				
				\vspace{3pt}
				call F-LCB to evaluate the individuals
				
				\vspace{3pt}
				selected solution $\bm{x}_p$ $\leftarrow$  EA
				
				\vspace{3pt}
				broadcast $\bm{w}$, $\bm{x}_p$ to clients $\in \bm{S}$
				
				\vspace{3pt}
				
			}
			
			\vspace{3pt}
		    $f(\bm{x}_p)$ $\leftarrow$ real evaluate $\bm{x}_p$ \quad /* not on the server*/
			\vspace{3pt}
			
			\For{{\bf Client} k $\in \bm{S}$ {\rm \textbf{ in parallel}}}
			{
			    \vspace{3pt}
		        /*update local surrogates*/
		    
			    \vspace{3pt}
                receive $\bm{w}$, $\bm{x}_p$ from the server
                
                \vspace{3pt}
				synchronize $\bm{w}_k \leftarrow \bm{w}$ 
				
                \vspace{3pt}
				
				add $\{\bm{x}_p, f_k(\bm{x}_p)\}$ to $D_k$
				\vspace{3pt}
				
				train $\bm{w}_k$ incrementally using $D_k^{init}\cup D_k$
				\vspace{3pt}
				
				upload $\bm{w}_k$ to server
				\vspace{3pt}
			}
			\vspace{3pt}

            $FE += 1$    \quad\quad\quad\quad    /* $\lambda N$ for distinct $\bm{x}_p$*/
			\vspace{3pt}
				
		}  

		\end{algorithm}
	\end{figure}

Note that in FDD-EA, sampling a candidate solution $\bm{x}_p$ on multiple clients at the same time is seen as one fitness evaluation (FE) in comparing with the centralized data-driven optimization. We consider this is fair since the evaluations on multiple clients are done in parallel, and only one same point is sampled. Also, only the participating clients receive the selected solution $\bm{x}_p$ and are involved in the next round of model updates. 

In the following, we will present in detail the proposed sorted model aggregation method, as well as the federated surrogate management strategy.   
\subsection{Sorted Averaging}
 \label{sec:sorted_ave}
In FDD-EA, training of the global surrogate is meant to effectively guide the evolutionary search instead of achieving accurate predictions. For this purpose, a federated surrogate management strategy is proposed to sample a new data on the participating clients in each round of model update, which is added to the existing database for training the local surrogate. Consequently, federated surrogate training in FDD-EA is an incremental learning process. 
	
	
In the vanilla federated learning, the training data on the local clients is stationary and a weighted averaging aggregation method, known as FedAvg \cite{mcmahan2016communication}, is proposed to parameter-wise average the uploaded local models according to the amount of the local data. Many variants have been proposed to enhance the learning performance in the presence of non-iid data and asynchronous model update \cite{chen2020twal}. However, since RBFNs are used for the surrogates, and the centers of the radial-basis-functions might have been shifted differently during the training. Thus, it has been found that a significant performance drop will occur if we average the centers, widths, and the weights of the RBFNs simply according to the index of the nodes.  

    \begin{figure}[htb!]
		\centering
		\includegraphics[width=3.3in]{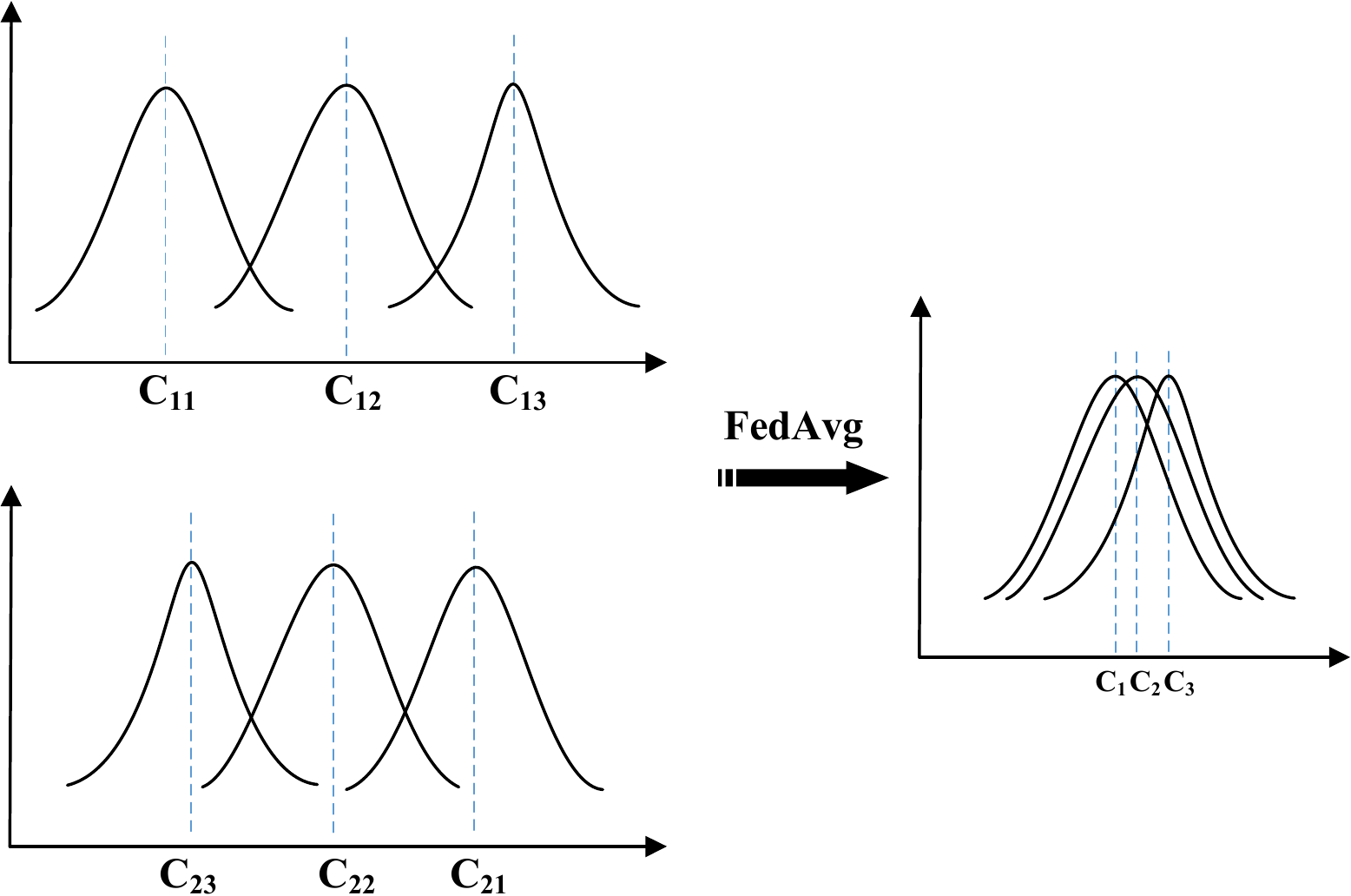}
		\caption{An illustrative example of aggregating two local univariate RBFNs, each having three nodes. The three Gaussian functions of the local RBFNs are shown on the left, and the resulting three Gaussian functions of the aggregated global RBFN are plotted on the right. Due to the shifted centers of the neurons, averaging the centers of the Gaussian functions according to the node index will lead to unreasonable results, causing possible serious performance degradation of the global surrogate.}
		\label{fig:mismatch}
	\end{figure}
    
After analysis, it is found that the serious performance degradation of the global model is caused by the averaging of very difference centers of the radial basis functions from different local models. Recall that the initial centers of the RBFNs are generated by clustering the decision variables of the training data. Given the same RBFN structure, the centers of the Gaussian functions of the corresponding nodes are similar, and a global surrogate can be generated by weighted averaging. However, it can happen after training, that the centers of the Gaussian functions have completely shifted, resulting in very different centers for the corresponding nodes of the RBFNs. Fig. \ref{fig:mismatch} provides an illustrative example of mismatches between two local RBFNs, where $C_{11}$, $C_{12}$ and $C_{13}$ and $C_{21}$, $C_{22}$ and $C_{23}$, respectively, represent the centers of the Gaussian functions of nodes $1$, $2$ and $3$ of the two RBFNs. If the Gaussian functions of the two RBFNs are averaged according to the index of the nodes, the resulting global RBFN will have three similar Gaussian functions with centers located on $C_1$, $C_2$ and $C_3$, respectively. As can be seen clearly on the right of Fig. \ref{fig:mismatch}, this global RBFN may perform very differently from the two local RBFNs, causing possible big performance drop.    
    
To alleviate this problem, we propose a sorted averaging method to obtain the global RBFN surrogate. The main idea of this strategy is to sort the centers of the radial basis functions of different models so that nodes with similar centers are averaged. To achieve this, let us denote $\bm{C}_k = (\bm{c}_{k,1}, \bm{c}_{k,2},..., \bm{c}_{k,m})$ as the center vectors of the $k$-th uploaded local surrogate $\bm{w}_k$, where $\bm{c}_{k,m} \in R^d$, $m$ is the number of centers. Then the following matching metric, which is the squared sum of the centers over all $d$ dimensions, is calculated for each of the $j$-th node ($j=1,\cdots,m$) of the $k$-th local RBFN:
    \begin{equation}
	\label{eq:center_distance}
	M_{k, \;j} = \sum_{i=1}^d c_{k, j,i}^2 \,, \,\,j=1,2,\cdots,m.
	\end{equation}
	
Therefore, the matching vector of the $k$-th RBFN $\bm{C}_k$ can be denoted by $\bm{M}_k = (M_{k, 1}, M_{k, 2},..., M_{k, m})$. Then, the index of nodes of the local RBFNs are sorted according to their matching vector in an ascending order. This way, all parameters, including the centers, widths, and the connecting weights are averaged according to the sorted index of the nodes of different local RBFNs: 
\begin{equation}
\bm{w} = \sum_{k=1}^{\lambda N} p_k \bm{w}_k,
\end{equation}
where $p_k$ is the weight as defined in Equation (\ref{eq:p_k}). The pseudo code of sorted averaging is summarized in Algorithm \ref{algo:sorted_average}.
	
	\begin{figure}[htb!]
		\removelatexerror
		\begin{algorithm}[H]
		
			\caption{Sorted Averaging}
			\label{algo:sorted_average}
			\SetKwInOut{Input}{Input}
			\SetKwInOut{Output}{Output}
			\Input{Local RBF surrogates $\bm{w}_k$ (contain centers $\bm{C}_k$, weights $\bm{a}_k$, spreads $\bm{\delta}_k$, bias $b_k$), global RBF surrogate $\bm{w}$.}
			\vspace{3pt}
			
			\text{{\bfseries Init:} Empty matching vector $\bm{M}_k$.}
			\vspace{3pt}
			
			\ForEach{k {\rm= 1, 2, 3 ... $\lambda N$}}
			{
			    \vspace{3pt}
				calculate $\bm{M}_k$ by equation (\ref{eq:center_distance})
				
				\vspace{3pt}
				index $\leftarrow$ sort($\bm{M}_k$)
				
				\vspace{3pt}
				sequence $\bm{C}_k$, $\bm{a}_k$, $\bm{\delta}_k$ by the index
				
				\vspace{3pt}
			     $\bm{w}_k = [\bm{C}_k; \; \bm{a}_k; \; \bm{\delta}_k; \; b_k]$
			}
			\vspace{3pt}
			
			$ \bm{w} \leftarrow \sum_{k=1}^{\lambda N}p_k \bm{w}_k$
			\vspace{6pt}
			
			\Output{ averaged $\bm{w}$}
		\end{algorithm}
	\end{figure}
    
\subsection{Federated Surrogate Management}
    \label{subsec:F-LCB}
In Bayesian optimization, the estimated uncertainty of the predictions provided by the Gaussian process plays an important role in the acquisition functions, since the uncertainty information is vital for striking a balance between exploration and exploitation. In the proposed FDD-EA, the global surrogate is built based on the weighted averaging of the local RBFNs and therefore the uncertainty of the the predictions needs to be properly estimated.   
	
In Section \ref{sec:framework}, we have mentioned that LCB instead of EI is adopted in our work. We prefer LCB over EI because the latter requires the current real minimums of the participating clients, leading to possible data privacy risks. To avoid this problem, we propose a federated LCB (F-LCB for short) as the acquisition function. The main idea is to make use of the predictions of both the global and local surrogates as well: 
    \begin{equation}
	\label{eq:modified_mean}
	\hat f(\bm{x}_p) = \frac{\hat f_\text{local}(\bm{x}_p)+\hat f_\text{fed}(\bm{x}_p)}{2},
	\end{equation}
where $\hat f_\text{local}(\bm{x}_p)$ is calculated by
	\begin{equation}
	\label{eq:local_mean}
	\hat f_\text{local}(\bm{x}_p) = \sum_k^{\lambda N} p_k \hat f_k(\bm{x}_p),
	\end{equation}
in which $p_k$ is the weight for the $k$-th local surrogate, $n_k$ is the number of data on the $k$-th client, Equation (\ref{eq:local_mean}) denotes the mean predicted value on $\bm{x}_p$ of all participated clients. And $f_\text{fed}(\bm{x}_p)$ is the predicted value of the federated global model, 
\begin{equation}
	\label{eq:f_fed}
	\hat f_\text{fed}(\bm{x}_p) = \bm{w}({\bm{x}_p}).
	\end{equation}
Consequently, the square of the standard deviation can be calculated by:
	\begin{equation}
	\label{eq:modified_std}
	\begin{aligned}
	\hat s^2(\bm{x}_p) = \frac{1}{\lambda N} &
	\left[\sum_k^{\lambda N}\left(\hat f_k(\bm{x}_p) - \hat f(\bm{x}_p)\right)^2 \right.  \\
	&+ \left. \left( \hat f_\text{fed}(\bm{x}_p) - \hat f(\bm{x}_p) \right)^2  \right].
	\end{aligned}
	\end{equation}
	
Finally, the federated acquisition function, F-LCB, can be calculated by replacing $\hat f(\bm{x}_p)$ and $\hat s^2(\bm{x}_p)$ in Equation (\ref{eq:LCB}) with the predicted fitness in Equations (\ref{eq:modified_mean}) and the estimated standard deviation in (\ref{eq:modified_std}), respectively. 

The combination of the local and global predictions, instead of the aggregated global surrogate only, aims to further increase the prediction quality and the quality of the uncertainty estimation for surrogate management. A comparative study with discussions will be presented in Section \ref{subsec:acq-study}. 

\section{Simulation Studies}
\label{sec:4}
To verify the effectiveness of FDD-EA, we first compare it with several state-of-the-art centralized data-driven evolutionary optimization algorithms on widely used benchmarks. Furthermore, we examine the performance of FDD-EA on the benchmarks when the data distributed on the clients are noisy, or when the data is non-iid. Finally, we present a comparative study of the proposed acquisition function to its variants.

\subsection{Experimental setting}
\subsubsection{Compared algorithms}
To examine the performance of the proposed algorithm, we compare FDD-EA with four popular online data-driven surrogate-assisted evolutionary algorithms on five benchmark problems, Ellipsoid, Rosenbrock, Ackely, Rastrigin and Griewank (the reader is referred to \cite{liu2013gaussian} for details of the benchmarks), with different numbers of decision variables ($d$ =10, 20, 30). A summary of the main features of the benchmarks is listed in Table \ref{table:settings}. The algorithms under comparison include CAL-SAPSO \cite{wang2017committee}, GPEME \cite{liu2013gaussian}, SHPSO \cite{yu2018surrogate}, and SSLPSO \cite{yu2019generation}. Note that all these SAEAs assume that the data is centrally stored and there does not exist any federated data-driven evolutionary optimization algorithms that address the problem in this work, to the best of our knowledge.

\begin{enumerate}
\item CAL-SAPSO: An online SAEA with committee-based active learning strategy, which uses an ensemble surrogate consisting of a quadratic polynomial regression (PR) model, an RBFN and a simple Kriging model. CAL-SAPSO adopts a query by committee model management strategy on the basis of prediction quality and amount of uncertainty.
\item GPEME: A Kriging-based online SAEA with LCB as the acquisition function to select solutions for real objective evaluations.
\item SHPSO: An RBF-based online SAEA, which combines the standard PSO and social learning PSO for optimization and selection of the solutions to be sampled and then updates the RBF surrogate model. 
\item SSLPSO: A surrogate-assisted social learning particle swarm optimization method. 
\end{enumerate}
    
In this work, a real-coded genetic algorithm (RCGA) \cite{kumar1995real, deb2000efficient} is selected as the base optimizer that applies the simulated binary crossover, polynomial mutation and tournament selection. The RCGA will run for 100 generations to find the minimum of the F-LCB. All algorithms under comparison collect $5d$ data pairs using the real fitness evaluations (FEs) for building the surrogate before optimization starts, and the optimization ends when a total of $11d$ FEs is exhausted. 

\subsubsection{Data partitions}
To thoroughly investigate the performance of the proposed algorithm, we verify its optimization performance on three types of data distributions: IID, noisy, and non-iid. 
    \begin{enumerate}
	    \item \textit{IID}: All clients are able to sample any data point in the entire decision space, and the initial points $\bm{x}_i$ ($i = {1,2,\cdots,5d}$) on all clients are the same, which are sampled using the LHS method. Meanwhile, all clients are able to sample any solutions identified by minimizing the federated acquisition function during the optimization. 
	    \item \textit{Noisy environments}: The same setting as the above, except that the fitness evaluations on all clients are subject to noise. Detailed definition of the noise is given in Section \ref{subsec:noisy}.
	    \item \textit{Non-IID}: In real-world applications, it is likely that some clients are not able to sample all data points specified by the acquisition function due to, for instance, different operating conditions. To examine the performance of FDD-EA subject to this constraint, we conduct experiments in which some data points specified by the acquisition function by the server are not accessible to some clients, resulting in non-iid data similar to federated learning.
	\end{enumerate}
	
    Note, however, that even in the IID case, the training data on different clients may be slightly different, due to the fact that in each round of model update, only a small portion of the clients participate and sample new data. 
	
	\renewcommand{\arraystretch}{1.5}
	\begin{table}[H]
		
		
		\centering
		
		\caption{Test Problems.}
		
		\label{table:settings}

		\begin{tabular}{|c|c|c|c|c|}
			
			\hline
			
			Problem         & $d$              &Optimum   &Characteristics  \\

			\hline
			Ellipsoid     	 & 10, 20, 30      &0.0      &  Uni-modal             \\
			
			Ackley          & 10, 20, 30      &0.0      & Multi-modal            \\
			
			Rastrigin       & 10, 20, 30      &0.0      & Multi-modal            \\
			
			Griewank        & 10, 20, 30      &0.0      & Multi-modal            \\
			
			Rosenbrock      & 10, 20, 30      &0.0      & Multi-modal            \\

			\hline
			
		\end{tabular}
		
	\end{table}
	
	\subsubsection{Parameter settings}
	The parameter settings of FDD-EA are as follows:
	
	\begin{itemize}
	    \item Total number of clients : $N$ = 100.
	    \item Participating ratio: $\lambda$ = 0.1.
	    \item Number of local epochs: $E$ = 20.
	    \item Learning rate for clients: $\eta$ = 0.12.
	    \item Number of RBF centers: $m=2d+1$.
	    \item Number of generations of EA: $g$ = 100.
	\end{itemize}
	
	The basis function of the RBF models is the Gaussian function, the $m$ centers are determined by the k-means clustering algorithm, and the widths are calculated by the maximum distance between the centers, the weights and the biases are trained using the gradient based method for $E=20$ epochs with a learning rate $\eta$ = 0.12. Each experiment is performed for 20 independent runs. 
    
    \subsection{Results on IID Data}
    \label{subsec:iid_bench}
	 
	\begin{table*}[htb!]
	\caption{Average best fitness values (shown as avg $\pm$ std) obtained by FDD-EA, CAL-SAPSO, GPEME, SHPSO and SSLPSO. The average ranks are obtained according to the Friedman’s test with the $p$-values being adjusted according to the Hommel’s procedure and the significance level of 0.05. FDD-EA is the control method.}
		
	\label{table:iid-results}
	\centering
        \begin{tabular}{|c|c|c|c|c|c|c|}
        \hline
        Problem    & $d$     & FDD-EA & CAL-SAPSO & GPEME & SHPSO & SSLPSO \\ \hline
        
        \multirow{3}{*}{Ellipsoid}
        & 10   & 6.17e-01 $\pm$ 3.13e-01  & \textbf{1.25e-01 $\pm$ 1.84e-01}  & 3.58e+01 $\pm$ 2.11e+01 & 3.34e+00 $\pm$ 1.92e+00 & 2.07e+00 $\pm$ 2.02e+00\\
        \cline{2-7}
        & 20   & \textbf{1.24e+00 $\pm$ 4.64e-01} & 2.32e+00 $\pm$ 1.14e-01  & 3.09e+02 $\pm$ 1.14e+02 & 1.34e+01 $\pm$ 4.79e+00 & 2.67e+01 $\pm$ 7.87e+00 \\
        \cline{2-7}
        & 30   &\textbf{3.23e+00 $\pm$ 4.18e-01}  & 3.10e+00 $\pm$ 2.09e+00  & 9.67e+02 $\pm$ 2.21e+02 & 4.87e+01 $\pm$ 1.76e+01 & 9.77e+01 $\pm$ 2.75e+01 \\
        \hline
        \multirow{3}{*}{Rosenbrock} 
        & 10   & \textbf{1.18e+01 $\pm$ 1.69e+00}  & 1.80e+01 $\pm$ 6.33e+00 & 1.67e+02 $\pm$ 9.30e+01 &9.20e+01 $\pm$ 5.99e+01 &3.08e+01 $\pm$ 1.01e+01\\ 
        \cline{2-7}
        & 20   & \textbf{2.22e+01 $\pm$ 1.18e+00} & 3.97e+01 $\pm$ 9.36e+00 & 9.30e+02 $\pm$ 4.19e+02 &1.79e+02 $\pm$ 7.35e+01 & 1.12e+02 $\pm$ 2.82e+01 \\ 
        \cline{2-7}
        & 30   & \textbf{3.55e+01 $\pm$ 1.05e+00} & 5.42e+01 $\pm$ 1.26e+01  & 2.04e+03 $\pm$ 9.09e+02 &2.12e+02 $\pm$ 5.49e+01& 2.46e+02 $\pm$ 5.85e+01 \\ 
        \hline
        \multirow{3}{*}{Ackley}
        & 10    & \textbf{4.36e+00 $\pm$ 3.36e-01} & 1.88e+01 $\pm$ 7.53e-01 & 1.64e+01 $\pm$ 2.88e+00 & 1.11e+01 $\pm$ 2.08e+00 & 7.61e+00 $\pm$ 1.24e+00 \\
        \cline{2-7}
        & 20   & \textbf{3.59e+00 $\pm$ 3.24e-01} & 1.72e+01 $\pm$ 2.90e+00  & 1.80e+01 $\pm$ 1.62e+00 & 1.12e+01 $\pm$ 1.95e+00 & 1.04e+01 $\pm$ 6.81e-01 \\
        \cline{2-7}
        & 30   & \textbf{4.41e+00 $\pm$ 3.70e-01} & 1.45e+01 $\pm$ 2.62e-01 & 1.83e+01 $\pm$ 1.16e+00 & 1.11e+01 $\pm$ 9.36e-01& 1.21e+01 $\pm$ 7.55e-01 \\ 
        \hline
        \multirow{3}{*}{Rastrigin}
        & 10   & \textbf{2.15e+01 $\pm$ 6.80e+00}  & 7.98e+01 $\pm$ 3.11e+01 & 5.99e+01 $\pm$ 1.59e+01 & 8.44e+01 $\pm$ 1.42e+01 & 6.85e+01 $\pm$ 1.43e+01\\ 
        \cline{2-7}
        & 20   & \textbf{3.32e+01 $\pm$ 6.56e+00}  & 7.58e+01 $\pm$ 2.07e+01  & 1.68e+02 $\pm$ 3.71e+01 & 1.76e+02 $\pm$ 2.07e+01 & 1.70e+02 $\pm$ 1.26e+01\\
        \cline{2-7}
        & 30   & \textbf{7.42e+01 $\pm$ 1.86e+01} & 8.28e+01 $\pm$ 1.39e+01 & 2.61e+02 $\pm$ 3.75e+01 & 2.76e+02 $\pm$ 2.43e+01 & 2.71e+02 $\pm$ 1.56e+01 \\ 
        \hline
        \multirow{3}{*}{Griewank}
        & 10   & 1.35e+00 $\pm$ 1.94e-01  & \textbf{1.22e+00 $\pm$ 1.64e-01}  & 2.89e+01 $\pm$ 2.25e+01 & 1.28e+00 $\pm$ 4.36e-01 & 2.29e+00 $\pm$ 4.95e-01 \\
        \cline{2-7}
        & 20   & 1.35e+00 $\pm$ 1.42e-01 & 1.37e+00 $\pm$ 1.40e-01  & 1.19e+02 $\pm$ 4.37e+01 & \textbf{1.15e+00 $\pm$ 1.23e-01} & 8.07e+00 $\pm$ 1.46e+00 \\ 
        \cline{2-7}
        & 30    & 1.50e+00 $\pm$ 1.82e-01  & 1.46e+00 $\pm$ 1.65e-01  & 2.79e+02 $\pm$ 6.46e+01  & \textbf{1.14e+00 $\pm$ 5.45e-02} & 1.84e+01 $\pm$ 2.97e+00 \\ 
        \hline
        \multicolumn{2}{|c|}{Average Rank} & 1.400 & 2.533   & 4.467 & 3.200 & 3.400  \\ \hline
        \multicolumn{2}{|c|}{Adjusted $p$-value}&NA &\textbf{0.050} &\textbf{0.000} &\textbf{0.004} &\textbf{0.002}  \\ \hline
        \end{tabular}
    \end{table*}
    
    \begin{figure*}[htb!]
		\centering
		\includegraphics[width=7in]{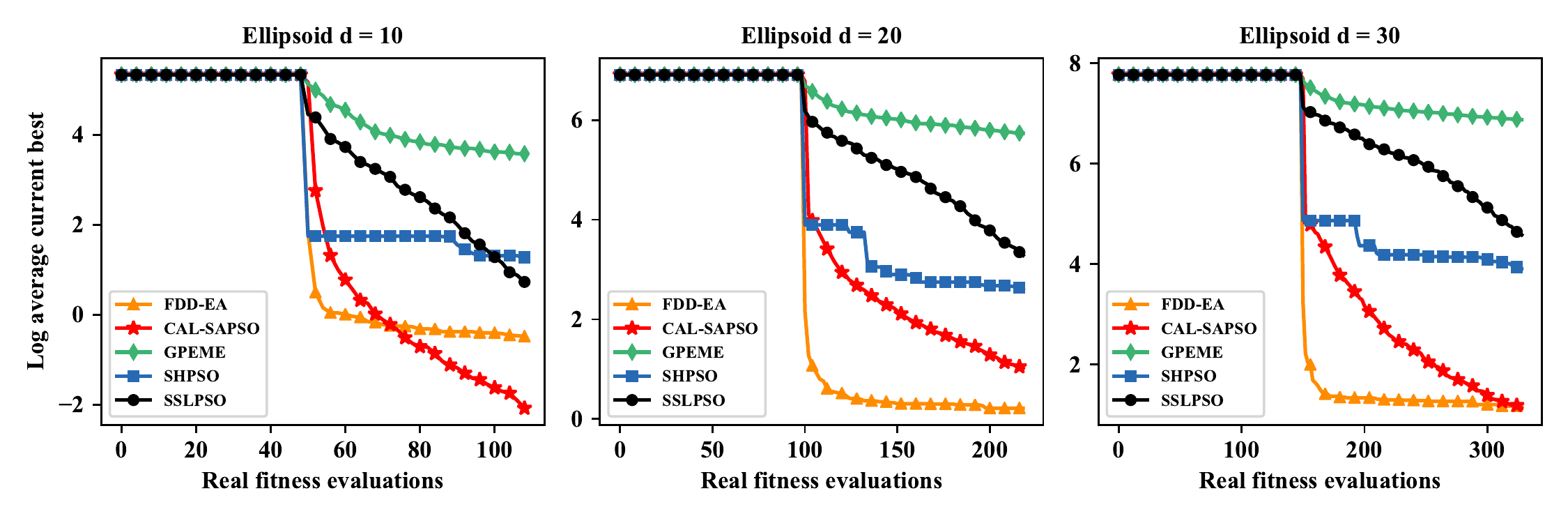}
		\caption{Convergence profiles of FDD-EA, CAL-SAPSO, GPEME, SHPSO and SSLPSO on the Elliposid function for $d=10,20,30$ in terms of the natural logarithm.}
		\label{fig:E_profiles}
	\end{figure*}
    
In the case of IID data distribution, LHS is used to sample the initial $5d$ data pairs on all clients. The results are presented in Table \ref{table:iid-results}, in which the average ranks of each algorithm are calculated by the Friedman’s test, and the $p$-values are adjusted according to the Hommel’s procedure \cite{derrac2011practical} with a significance level of 0.05. The better results are highlighted. It can be concluded that the proposed FDD-EA performs significantly better than the compared algorithm on 11 out of 15 instances. This indicates that FDD-EA performs competitively in comparison with the state-of-the-art centralized data-driven evolutionary algorithms. 
	
	\begin{figure*}[htb!]
		\centering
		\includegraphics[width=7in]{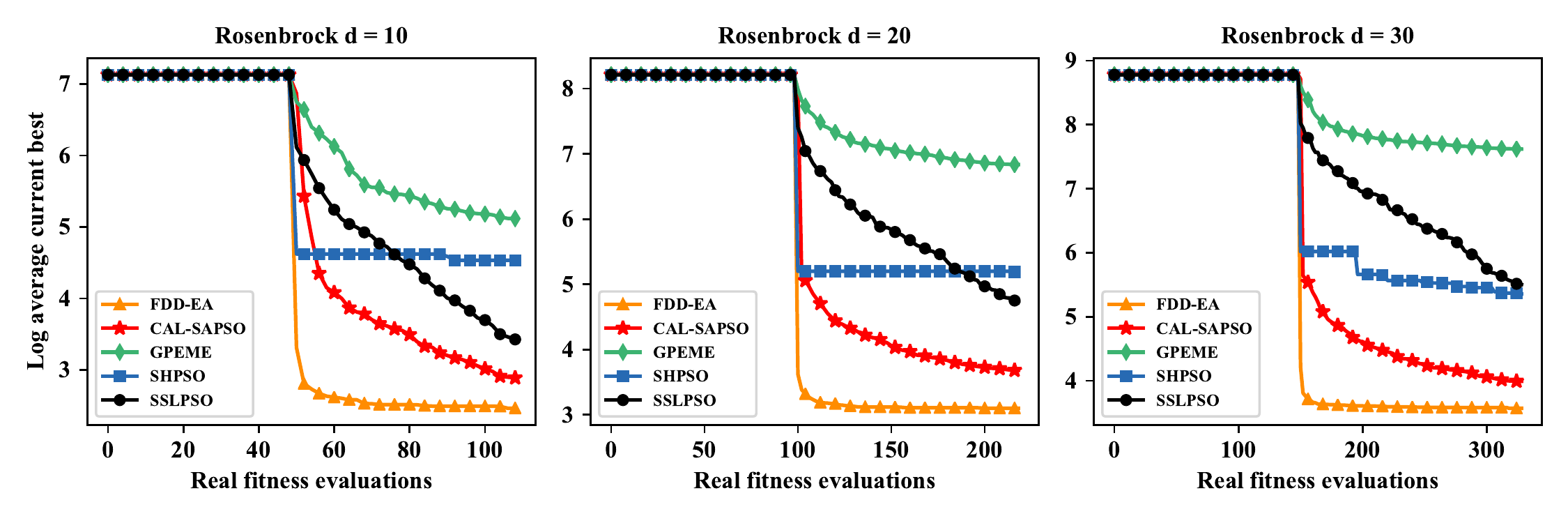}
		\caption{Convergence profiles of FDD-EA, CAL-SAPSO, GPEME, SHPSO and SSLPSO on the Rosenbrock function when $d=10,20,30$  in terms of the natural logarithm.}
		\label{fig:Ros_profiles}
	\end{figure*}
	
	\begin{figure*}[htb!]
		\centering
		\includegraphics[width=7in]{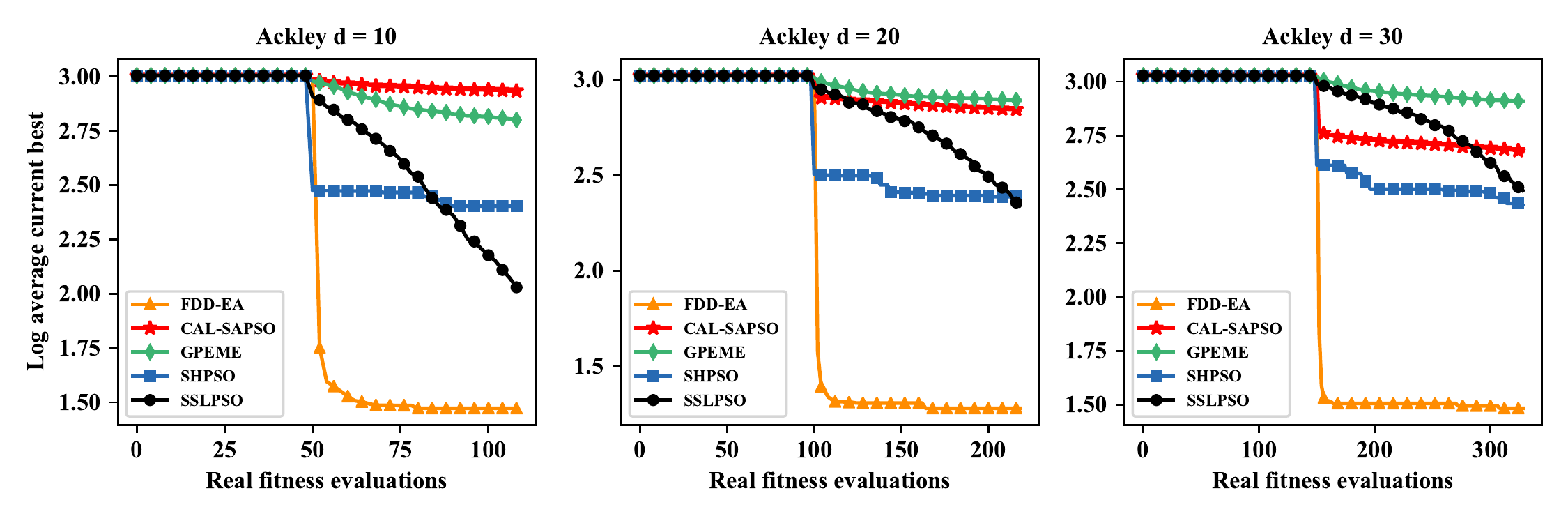}
		\caption{Convergence profiles of FDD-EA, CAL-SAPSO, GPEME, SHPSO and SSLPSO on the Ackley function when $d=10,20,30$.}
		\label{fig:A_profiles}
	\end{figure*}
	
	\begin{figure*}[htb!]
		\centering
		\includegraphics[width=7in]{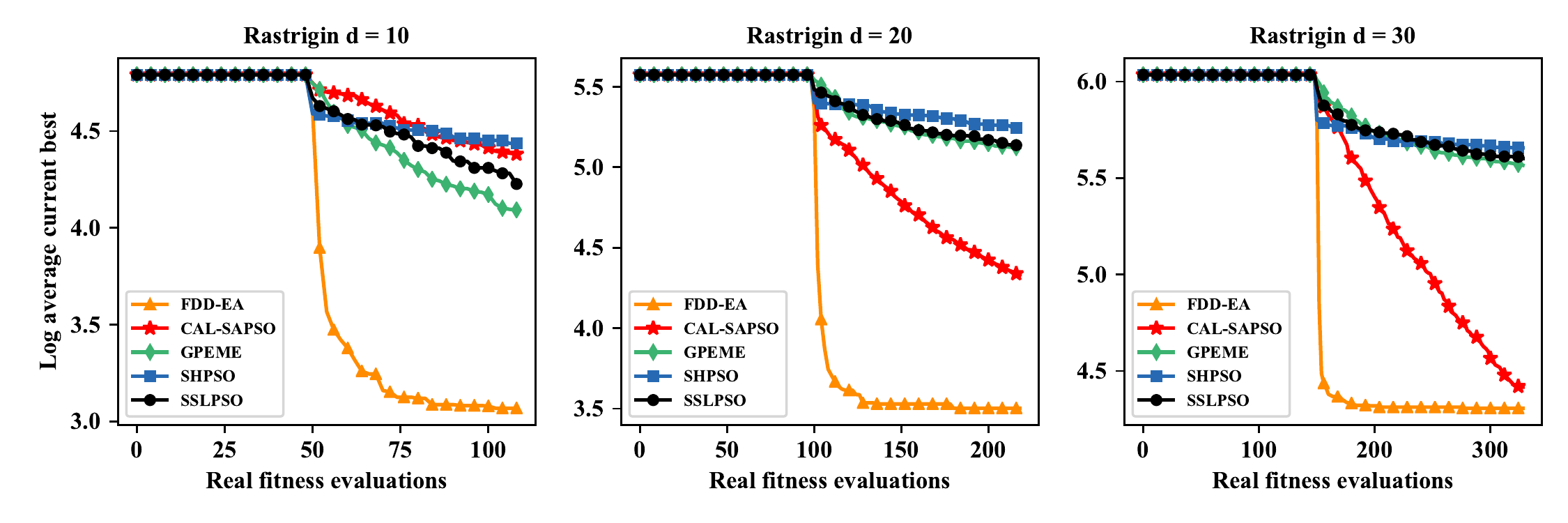}
		\caption{Convergence profiles of FDD-EA, CAL-SAPSO, GPEME, SHPSO and SSLPSO on the Rastrigin function when $d=10,20,30$ in terms of the natural logarithm.}
		\label{fig:Rast_profiles}
	\end{figure*}
	
	\begin{figure*}[htb!]
		\centering
		\includegraphics[width=7in]{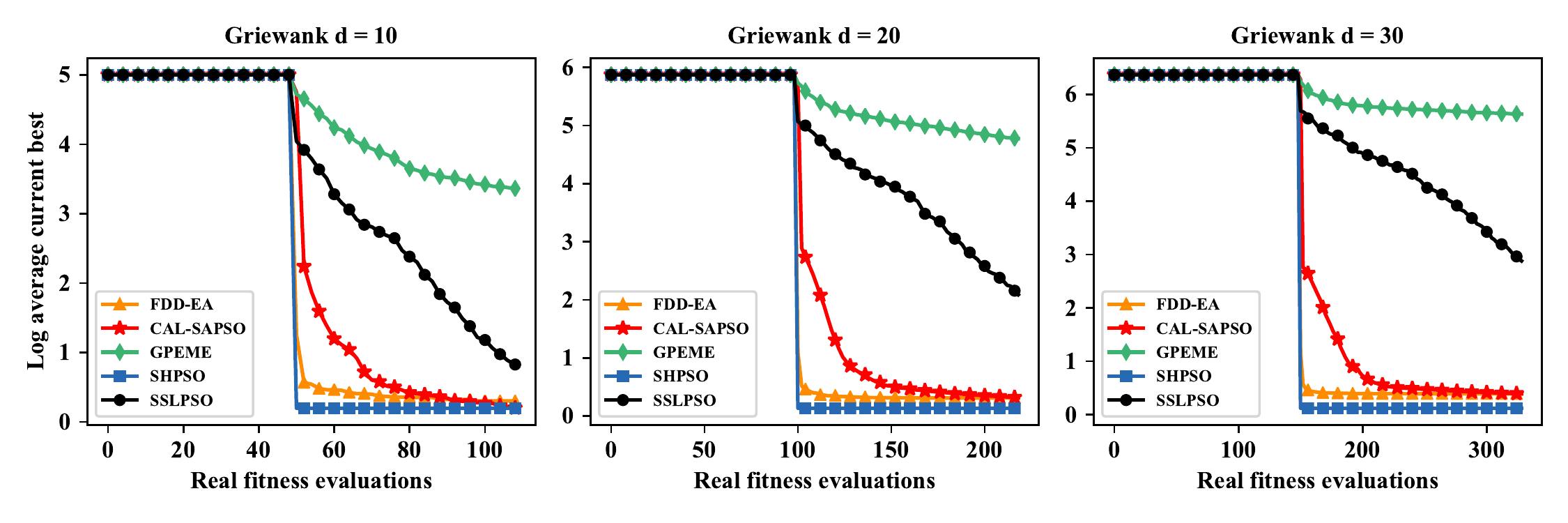}
		\caption{Convergence profiles of FDD-EA, CAL-SAPSO, GPEME, SHPSO and SSLPSO on the Griewank function when $d=10,20,30$ in terms of the natural logarithm.}
		\label{fig:G_profiles}
	\end{figure*}
	
To take a closer look at the performances of the compared algorithms, the convergence profiles of the compared algorithms on the 10D, 20D and 30D test functions are presented in Fig. \ref{fig:E_profiles}, \ref{fig:Ros_profiles}, \ref{fig:A_profiles}, \ref{fig:Rast_profiles} and \ref{fig:G_profiles}, respectively. From these results, we can observe that FDD-EA clearly outperforms all other compared algorithms on the Rosenbrock, Ackley and Rastrigin functions. On the Elliposid function, CAL-SAPSO and FDD-EA outperforms other algorithms under comparison, although FDD-EA performs worse than CAL-SAPSO on the 10D instance. On the Griewank function, FDD-EA is outperformed by CAL-SAPSO or SHPSO, nevertheless, still clearly outperforms SSLPSO and GPEME.

To summarize, FDD-EA performs the best on 11 out of 15 instances in the IID data environment, and it consistently converges fast on all test instances. From these results we can conclude that FDD-EA and the ensemble-based method CAL-SAPSO are more competitive than the other compared algorithms. However, it is noticed that FDD-EA quickly gets stuck in a local optimum in the later search stage. 
	
	
We also compare FDD-EA with SA-COSO on high-dimensional problems up to 100D, because the online centralized SAEAs compared above are designated for optimization problems up to 30 decision variables. Therefore, we compare FDD-EA with the surrogate-assisted co-operative swarm optimization algorithm (SA-COSO) \cite{sun2017surrogate} designed for high-dimensional expensive optimization on the five test instances of 50D and 100D, respectively. The experimental results are presented in Table \ref{Tab.S1} and Figs. \ref{fig:E50_100}, \ref{fig:Ro50_100}, \ref{fig:A50_100}, \ref{fig:Ra50_100}, \ref{fig:G50_100} of the Supplementary material in the Appendix \ref{appendix:A}. From these results, we can see that FDD-EA outperforms SA-COSO on all instances except for the 100D Griewank function.
    
\subsection{Noisy Fitness Evaluations}
\label{subsec:noisy}
In data driven optimization, data might be collected from production process, which is subject to noise \cite{guo2016} \cite{chugh2017}. To study robustness of the proposed algorithm against noisy fitness evaluations, we add 
\begin{equation}
\label{eq:noisy_benchmark}
f_k(\bm{x}) = f(\bm{x})+\alpha \,\xi, \;\forall \; k \in N
\end{equation}
where $\alpha \in [0,1]$ is a constant that determines the magnitude of the noise, $\xi$ is noise sampled from the standard Gaussian distribution $N(0, 1)$, and $f(\bm{x})$ denotes the benchmark problem.
   
\begin{figure}[htb!]
		\centering
		\includegraphics[width=3.4in]{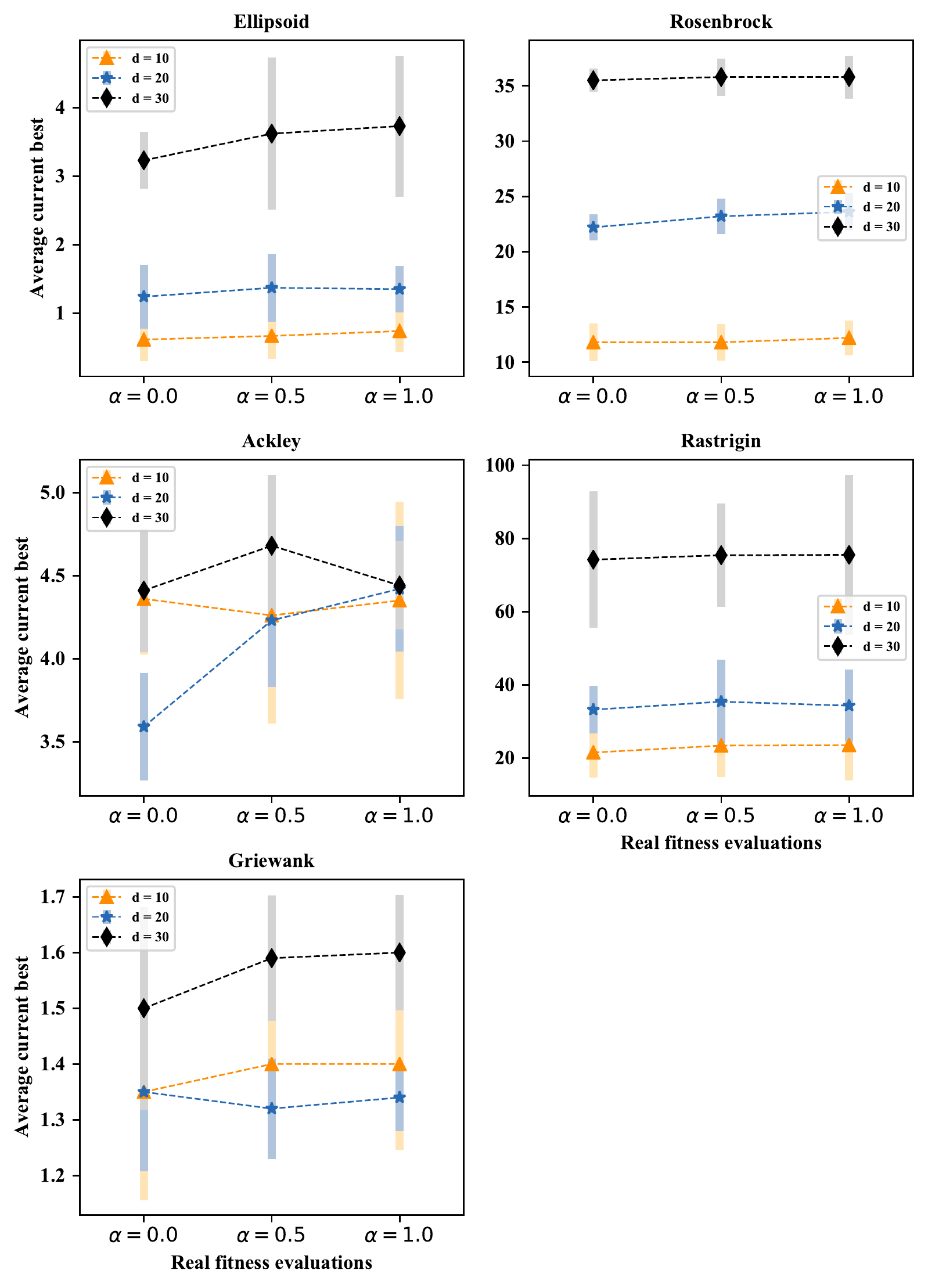}
		\caption{The mean fitness and the standard deviation of FDD-EA on 15 test instances in the presence of noisy fitness evaluations when the noise level $\alpha$ changes from 0 to 1.}
		\label{fig:noisy}
	\end{figure}

We examine the performance of FDD-EA on the above-mentioned benchmark problems, with all settings being the same as in Section \ref{subsec:iid_bench} and present the results in Fig. \ref{fig:noisy}. From these results, we find that the performance of FDD-EA on all the tested instances except for the 20D Ackley and 30D Griewank functions does not seriously deteriorate when the noise level changes from 0 to 1. Thus, we conclude that FDD-EA is fairly insensitive to noise in the fitness evaluations.   
	
\subsection{Non-IID}
\label{subsec:noniid}
Here, we examine the performance of FDD-EA on non-IID data distribution, considering the fact that different clients may have different operating conditions and specifications, which leads to the situation where some regions in the decision space become infeasible on some local clients. To take this situation into account, we introduce an infeasible domain $dom(k, \tau)$ for the $k$-th client to determine whether a candidate point $\bm{x}_p$ can be sampled by the $k$-th client.
    
\begin{equation}
	\label{eq:noniid_condition}
	\begin{aligned}
	& dom(k, \tau) = \\
	&\left[ x_{lb}+(k-1) g_k, \;\min\{x_{lb}+(k+\tau-1) g_k, \;x_{ub}\} \right], \\
	\end{aligned}
	\end{equation}
where $x_{lb}$, $x_{ub}$ are the lower and upper bounds of one certain dimension of $\bm{x}$, respectively, $\tau$ is an integer control parameter that determines the range of the infeasible domain, and $g_k$ is calculated by 
\begin{equation}
	\label{eq:noniid_gap}
	g_k = \frac{x_{ub}-x_{lb}}{N}
\end{equation}
	
Specifically, if the value of a certain dimension of a sample does not fall in the infeasible domain $dom(k, \tau)$, $\bm{x}_p$ will be sampled by the $k$-th client; otherwise, the client fails to sample $\bm{x}_p$ and no new data sample will be generated. Note that we have the IID case for $\tau = 0$. The sampling constraint imposed in (\ref{eq:noniid_condition}) will lead to more different distributions of the newly sampled data on different clients, making the federated optimization more challenging. The same settings as in the experiment on IID data are adopted, except that data sampling is now constrained. Again, all results are averaged over 20 independent runs.

\begin{figure}[htb!]
		\centering
		\includegraphics[width=3.4in]{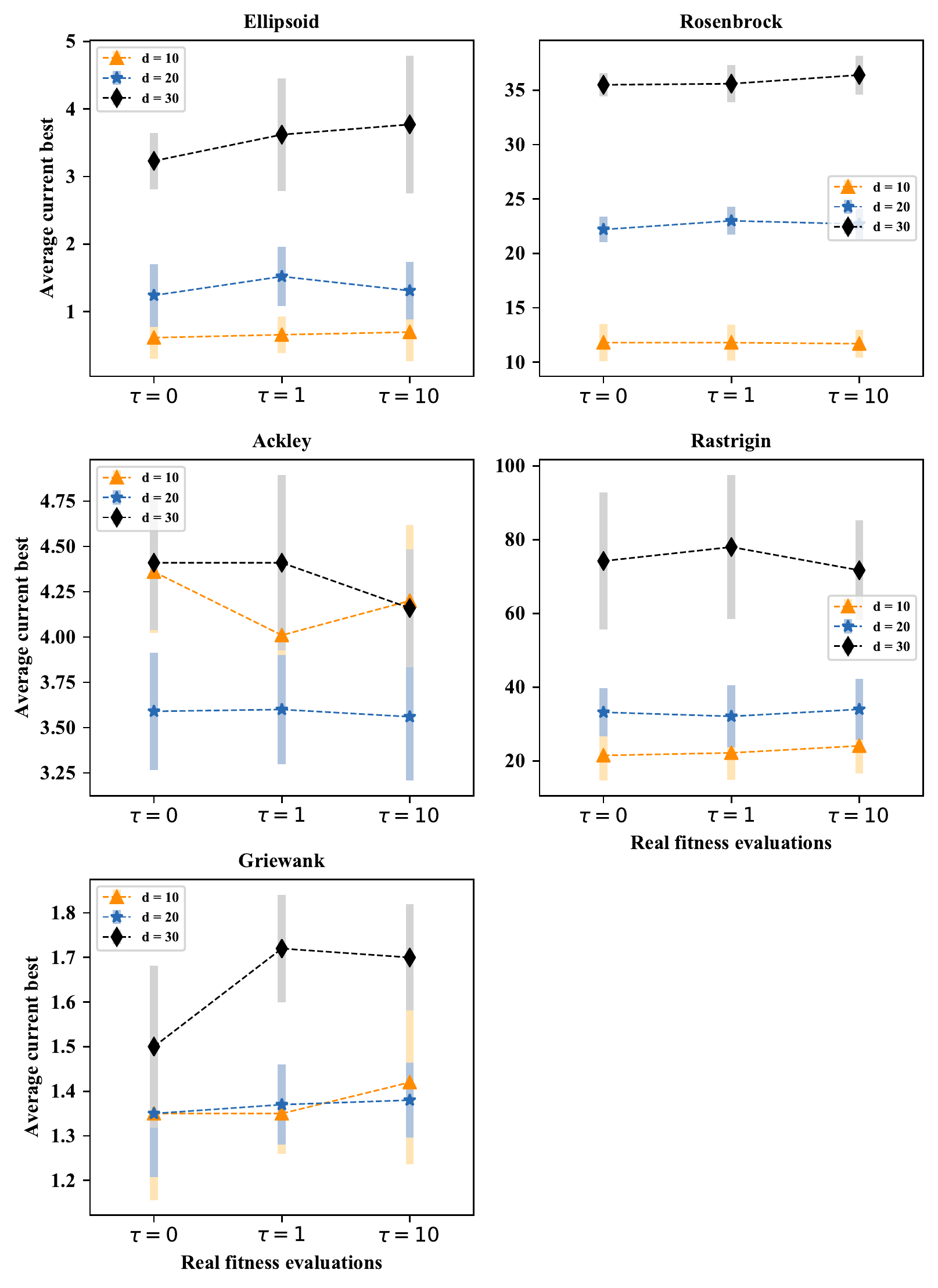}
		\caption{The convergence profiles of FDD-EA on non-IID data with an infeasible domain under different $\tau$ values.}
		\label{fig:infeasible_table}
\end{figure}

The average optimization results together with their standard deviations over 20 independent runs are presented in Fig. \ref{fig:infeasible_table}. From these results, we can conclude that FDD-EA is robust against the non-IID of the newly sampled data, although the performance has a slight degradation on a few instances, such as on the 30D Ellipsoid and Griewank functions. Note also that the performance of FD-FO even slightly enhances, for instance, on the 10D and 30D Ackley functions. 

\begin{figure}[htb!]
		\centering
		\includegraphics[width=3.4in]{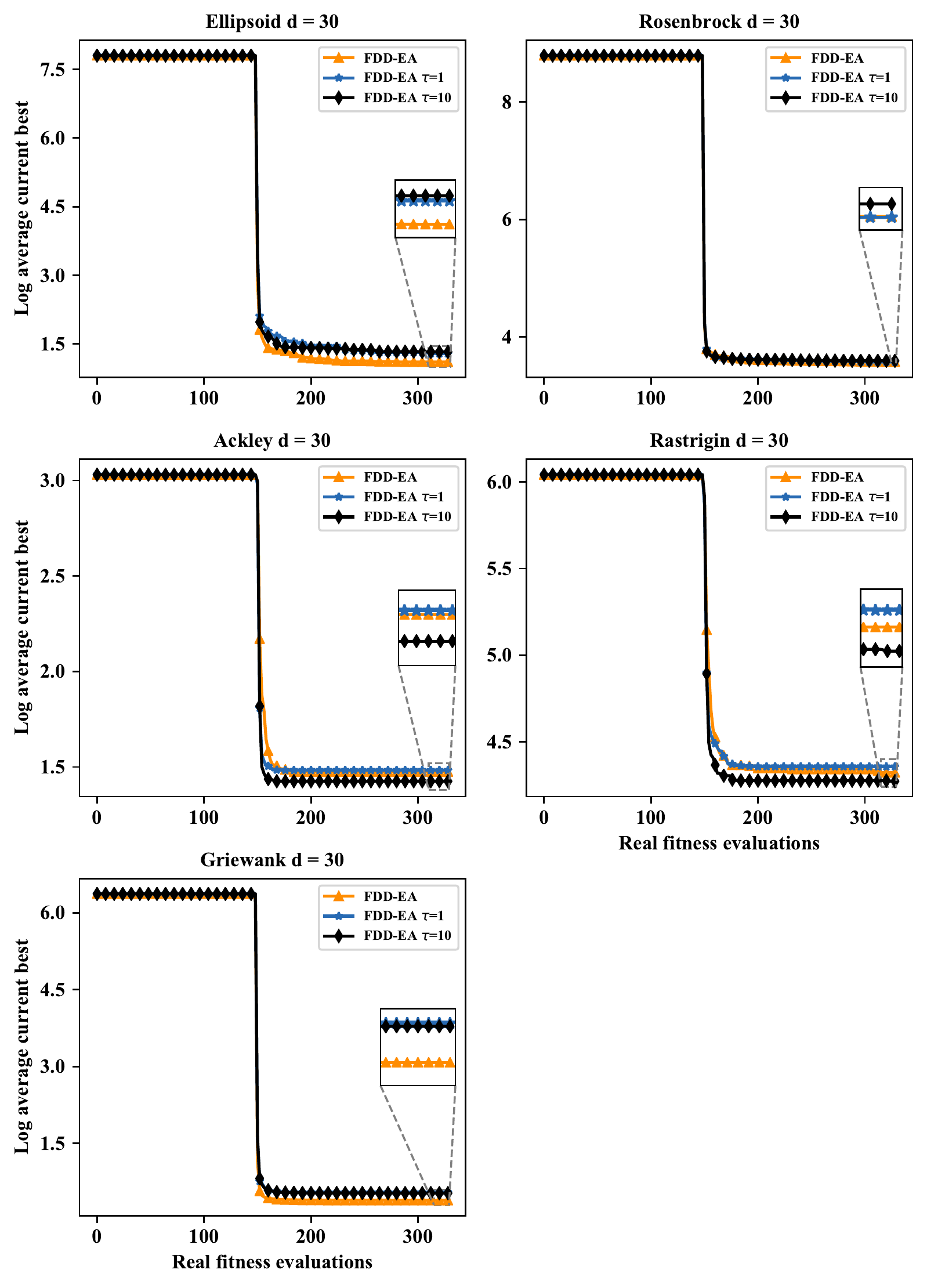}
		\caption{The convergence profiles of non-IID with an infeasible domain for different $\tau$ values.}
		\label{fig:infeasible}
\end{figure}

For a further investigation of the performance change, we plot the convergence profiles of FDD-EA on the five 30D benchmark functions in Fig. \ref{fig:infeasible}. Note that the size of real evaluated samples decreases as $\tau$ increases, resulting in a slight deterioration in the performance of FDD-EA. 

\subsection{Comparison of Surrogate Management Strategies}
	\label{subsec:acq-study}
To demonstrate the efficiency of the proposed federated surrogate management strategy (F-LCB), we compare it with its two variants: L-LCB calculates the mean fitness based on the weighted average of the predictions of the local surrogates using Equation (\ref{eq:local_mean}), while G-LCB predicts the mean fitness using the global surrogate according to Equation (\ref{eq:f_fed}). Accordingly, the uncertainty of the predictions for L-LCB and G-LCB  are calculated as follows:
	\begin{equation}
	\label{eq:local_std}
	\begin{aligned}
	\hat s^2(\bm{x}_p) = \frac{1}{\lambda N-1} &
	\left[\sum_k^{\lambda N}\left(\hat f_k(\bm{x}_p) - \hat f(\bm{x}_p)\right)^2 \right].  \\
	\end{aligned}
	\end{equation}

	\begin{table}[htb!]
    	\caption{Average best fitness values (shown as avg $\pm$ std) of FDD-EA and its two variants using a different acquisition function.}
    		
    	\label{table:F-LCB}
    	\centering
        \begin{tabular}{|c|c|c|c|c|}
        \hline
        Problem    & $d$ &F-LCB & L-LCB                & G-LCB              \\ \hline
        \multirow{3}{*}{Ellipsoid} 
        & 10  & \textbf{0.62 $\pm$ 0.31} & 4.73 $\pm$ 1.94 & 1.07 $\pm$ 0.44 \\ \cline{2-5}
        & 20  & \textbf{1.24 $\pm$ 0.46} & 18.44 $\pm$ 10.51 & 2.71 $\pm$ 0.89 \\ 
        \cline{2-5}
        & 30  & \textbf{3.23 $\pm$ 0.42} & 19.22 $\pm$ 7.60 & 5.97 $\pm$ 1.14 \\ 
        \hline
        \multirow{3}{*}{Rosenbrock}
        & 10  & \textbf{11.75 $\pm$ 1.69} & 44.32 $\pm$ 14.34 & 16.10 $\pm$ 4.11 \\ 
        \cline{2-5}
        & 20  & \textbf{22.22 $\pm$ 1.18} & 78.11 $\pm$ 17.98 & 28.60 $\pm$ 2.43 \\ 
        \cline{2-5}
        & 30  & \textbf{35.54 $\pm$ 1.05} & 73.30 $\pm$ 24.0 & 40.73 $\pm$ 2.26 \\ 
        \hline
        \multirow{3}{*}{Ackley}
        & 10  & \textbf{4.36 $\pm$ 0.34} & 7.41 $\pm$ 1.61 & 4.42 $\pm$ 0.43 \\ 
        \cline{2-5}
        & 20  & \textbf{3.59 $\pm$ 0.32} & 6.80 $\pm$ 0.90 & 4.14 $\pm$ 0.38 \\ 
        \cline{2-5}
        & 30  & \textbf{4.41 $\pm$ 0.37} & 6.60 $\pm$ 0.56 & 5.01 $\pm$ 0.44 \\ 
        \hline
        \multirow{3}{*}{Rastrigin}  
        & 10  & \textbf{21.51 $\pm$ 6.80} & 59.14 $\pm$ 12.77 & 35.77 $\pm$ 9.98 \\ 
        \cline{2-5}
        & 20  & \textbf{33.19 $\pm$ 6.56} & 138.68 $\pm$ 18.60 & 66.91 $\pm$ 15.84 \\ 
        \cline{2-5}
        & 30  & \textbf{74.21 $\pm$ 18.6} & 199.77 $\pm$ 18.38 & 116.40 $\pm$ 25.83 \\ 
        \hline
        \multirow{3}{*}{Griewank} 
        & 10  & \textbf{1.35 $\pm$ 0.19} & 2.36 $\pm$ 0.34 & 1.60 $\pm$ 0.23 \\ 
        \cline{2-5}
        & 20  & \textbf{1.35 $\pm$ 0.14} & 1.76 $\pm$ 0.25 & 1.53 $\pm$ 0.15 \\
        \cline{2-5}
        & 30  & \textbf{1.50 $\pm$ 0.18} & 1.94 $\pm$ 0.18 & 1.83 $\pm$ 0.22 \\ 
        \hline
        \end{tabular}
    \end{table}
	
\begin{figure}[htb!]
		\centering
		\includegraphics[width=3.4in]{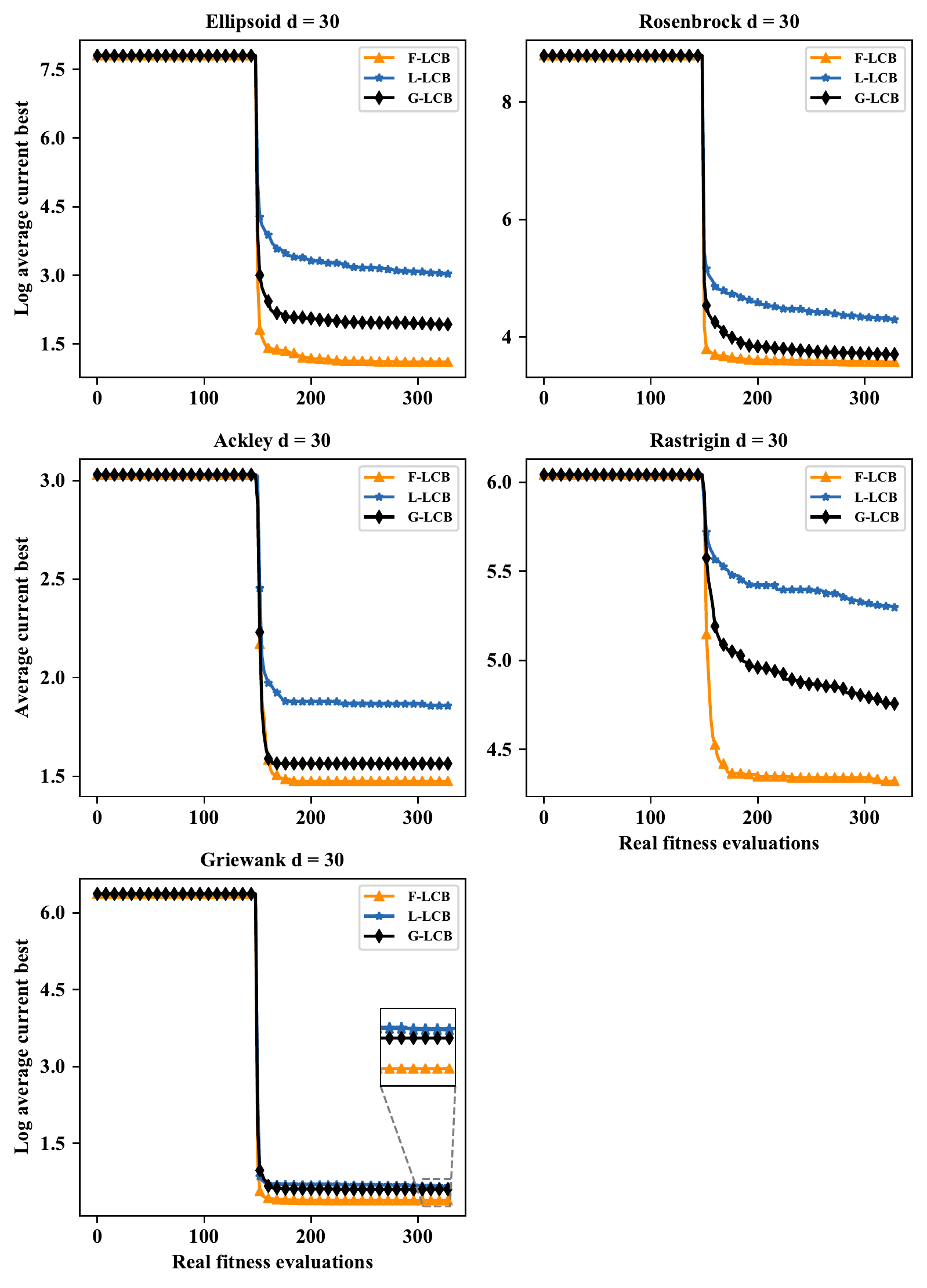}
		\caption{The convergence profiles of FDD-EA and its variants using a different acquisition function.}
		\label{fig:F-LCB}
	\end{figure}

The results averaged over 20 independent runs are presented in Table \ref{table:F-LCB} with the same parameter settings as in the experiments on the IID data except for the acquisition function. These results indicate that F-LCB consistently outperforms G-LCB and F-LCB on all the test instances studied in this work, and the advantage becomes even clearer when the dimension increases from 10 to 30. To further illustrate the performance differences resulting from the acquisition functions, the convergence profiles of F-LCB, L-LCB and G-LCB on the five 30D test functions are presented in Fig. \ref{fig:F-LCB}. One general observation we can make is that all methods converge very fast in the early search stage (recall that the first 150 FEs are offline samples for the 30D test instances), however, the optimization stagnates quickly after a small number of federated search steps. 
	
\subsection{Sensitivity Analysis}
To investigate the impacts of the local epoch $E$, the participation ratio $\lambda$ and the learning rate $\eta$ on the performance of the proposed FDD-EA, empirical studies are carried out on the 10D Ellipsoid function and the results are described in the Section \ref{Supp:A} of the Supplementary material in the Appendix \ref{appendix:A}. From Fig. \ref{fig:epoch} - \ref{fig:eta}, we can conclude that the performance of the proposed algorithm is relatively insensitive to these parameter settings. Interestingly, FDD-EA can achieve satisfactory results also for a small $E$, which is attractive for online applications of proposed method. We also set $E$ to 20 in all our empirical studies presented above. 
    
\section{Conclusions and Future Work}
	\label{sec:5}
    Existing data-driven evolutionary algorithms assume all data is centrally stored. In many real-world problems, by contrast, the data is collected on multiple sites and not allowed to be transmitted to a central server for security and privacy reasons, as well as for computational and real-time requirements. To bridge the gap, this paper proposes a federated data-driven evolutionary algorithm, FDD-EA, for distributed optimization. In FDD-EA, a sorted averaging method is designed for aggregating the local RBF surrogates to generate the global surrogate. In addition, a federated acquisition function is proposed to make use of the information in both local and the global surrogates. Our comparative studies on five benchmark problems demonstrate the effectiveness of the proposed FDD-EA, also in the presence of noisy and non-iid data.  
	
	However, several questions remain open. For example, although FDD-EA converges fast in the beginning of the search and is able to achieve competitive performance in comparison to existing centralized SAEAs, it can make only minor improvements after a certain number of search rounds, which might be attributed to the inefficiency of the present incremental learning algorithm in the federated optimization environment. In addition, the data distributions considered in this work is relatively ideal, taking into account the fact that in practice the data may be vertically partitioned, which will make the federated surrogate training and optimization much more challenging. Furthermore, it is desirable to extend the proposed framework to multi- and many-objective data-driven optimization problems. Finally, validation of the proposed framework on real-world problems will be our future target.  
	
	\appendices
	\section{Supplementary Material}
	\label{appendix:A}
	This is the supplementary material of the paper "A Federated Data-Driven Evolutionary Algorithm", providing some additional experimental results. In this material, we start by presenting the results of sensitivity analysis of three important parameters, the local learning epochs, participation ratio and larning rate, on the performance of the proposed algorithm. Finally, an comparison of FDD-EA with SA-COSO is conducted to examine the performance of the proposed algorithm on 50- and 100-dimensional optimization problems.
	
	\subsection{Sensitivity Analysis}
	\label{Supp:A}
	To investigate the impact of the important parameters of FDD-EA on its performance, we conduct sensitivity analysis of three parameters, namely the number of epochs $E$ in local training, participation ratio $\lambda$, and learning rate $\eta$. We present the results on the 10-D Ellipsoid function as an illustrative example, and similar conclusions can be drawn on other test instances. 
	
	\subsubsection{Sensitivity Analysis of Number of Local Epochs}
\begin{figure*}[htb!]
		\centering
		\includegraphics[width=0.9\textwidth]{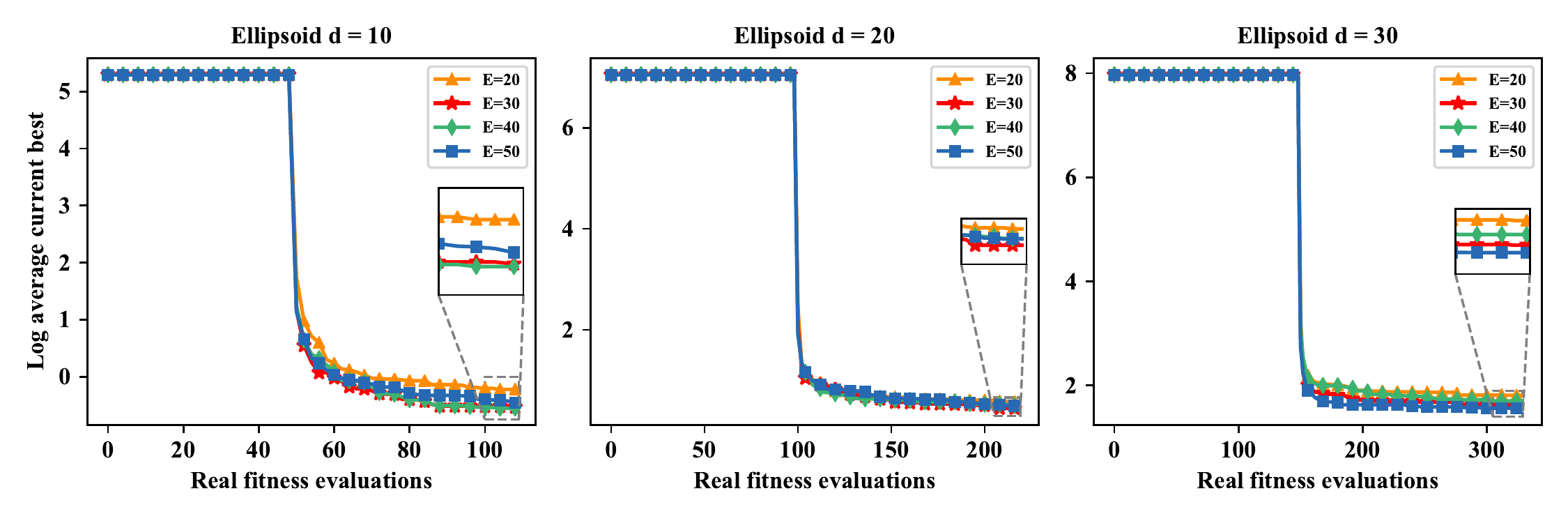}
		\caption{Convergence profiles of FDD-EA in terms of the natural logarithm on the 10D-30D Ellipsoid functions with different local epochs.}
		\label{fig:epoch}
	\end{figure*}

	To avoid the disturbance introduced by newly added samples, we look at the performance changes on the offline data only when $E$ is set to 20, 30, 40, and 50. The convergence profiles of FDD-EA on 10D, 20D and 30D Ellipsoid function averaged over 20 independent runs are illustrated in Fig. \ref{fig:epoch}. The performance fluctuates when $E$ changes, although the changes are minor. The best performance on the 10 is achieved when $E=20$. The performance drops more seriously when $E$ drops to 20, indicating that the training might be insufficient, As we can see, when $E=20$, the performance of FDD-EA is worse than others, increasing the local epoch to 30 and 40 lead to a better performance, which may be due to the underfitting problem when $E$ is small. However, in the case of $E \ge 40$, the increasement of $E$ will cause a slight performance degeneration, which may be caused by the overfitting problem. In this paper, wo have not adjust the local epoch $E$ frequently, and use $E=20$ for all experiments. The obtained satisfactory experimental results proves the robustness of FDD-EA on different local epochs and the positive effect of the server model.
	
	\subsubsection{Sensitivity Analysis of Participation Ratio}
\begin{figure*}[htb!]
		\centering
		\includegraphics[width=0.9\textwidth]{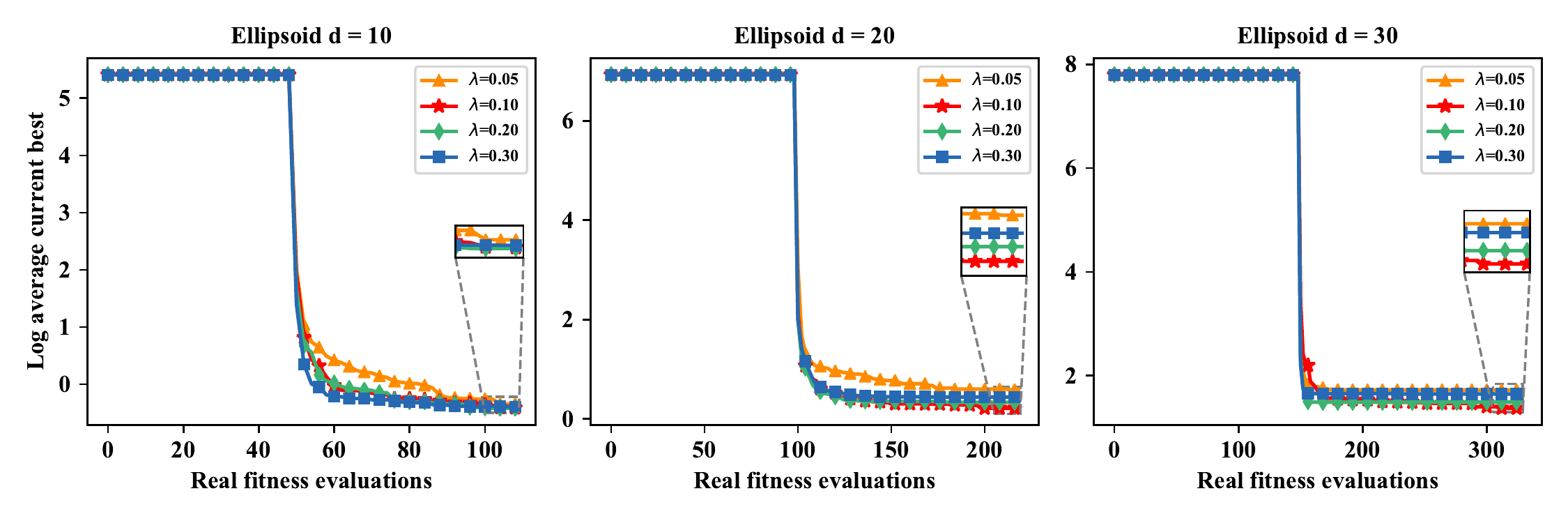}
		\caption{Convergence profiles of FDD-EA in terms of the natural logarithm on the 10D-30D Ellipsoid functions with different participation ratios.}
		\label{fig:ratio}
	\end{figure*}

	As we know, participation ratio $\lambda$ is a parameters of important practical significance. Intuitively, the participation ratio determines the number of participants, and in real-world applications, $\lambda$ is always influenced by the bandwidth, the remaining power of edge devices and the computational budgets. Generally, low participation rates make federated leanring more challening. Here, we set $\lambda = 0.05, 0.10, 0.20, 0.30$ to investigate its influence on the optimization results. The convergence profiles of FDD-EA on the 10D Ellipsoid function are illustrated in Fig. \ref{fig:ratio}. We see that the performance of FDD-EA is insensitive to the participation ratio, although a slight performance degeneration is observed when $\lambda=0.05$, meaning five clients participate in each round of updates.
	
	\subsubsection{Sensitivity Analysis of Learning Rate}
\begin{figure*}[htb!]
		\centering
		\includegraphics[width=0.9\textwidth]{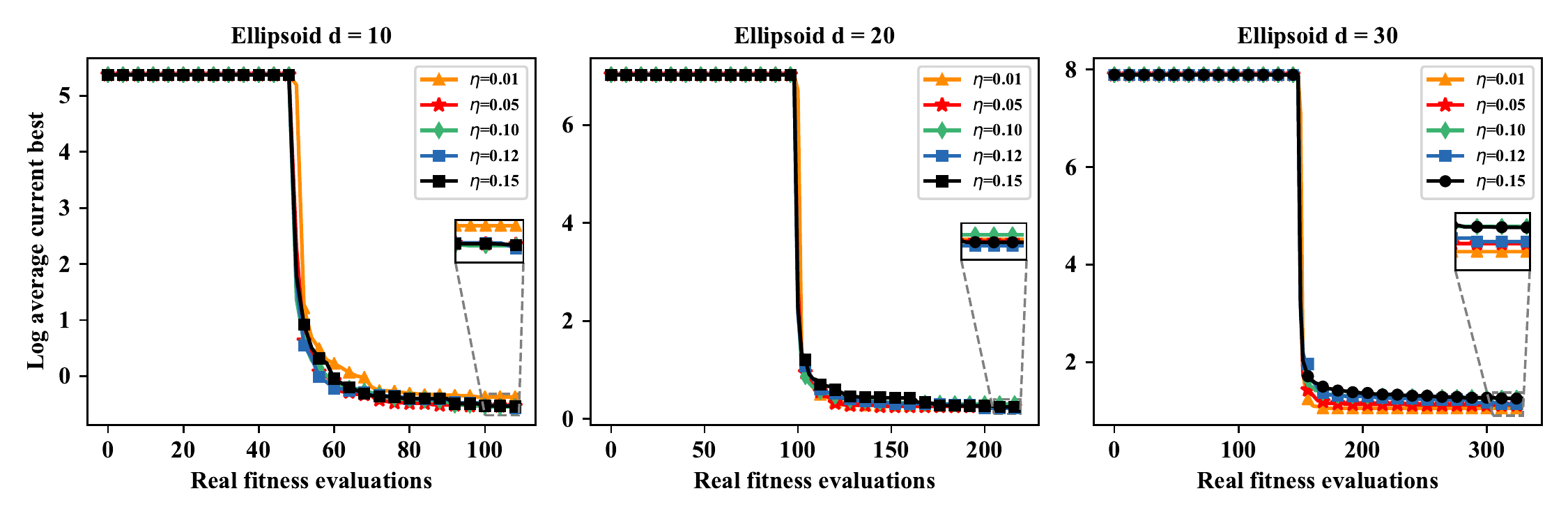}
		\caption{Convergence profiles of FDD-EA in terms of the natural logarithm on the 10D-30D Ellipsoid functions with different learning rates.}
		\label{fig:eta}
	\end{figure*}

	The training process of local RBFN models are based on gradient the descend method and hence, the learning rate is also an important factor. To invesitgate its influence, we set the learning rate $\eta$ to 0.01, 0.05, 0.10, 0.12, 0.15. The convergence profiles are plotted in Fig. \ref{fig:eta}. It can be concluded that FDD-EA performs well when $\eta = 0.05,0.10,0.12,0.15$. We also notice that the performance starts to drop when $\eta = 0.01$ and $d=10$, which imply that the training is inadequte. When $d=30$, the lower learning rates lead to slightly a better performance due to the enough training data and communication rounds. Hence, we can introduce learning rate decay for FDD-EA in high dimensional problems.
	
	\subsection{Results on 50- and 100-dimensional Benchmark Problems}
	In order to examine the ability of FDD-EA to deal with high-dimensional optimization problems, we further compare FDD-EA with one state-of-the-art SA-COSO \cite{sun2017surrogate} on 50- and 100-dimensional benchmark problems. The parameter settings are the same as the previous experiments. Both algorithms use $5d$ offline samples and a maximum of $11d$ real fitness evaluations. $2d+1$ nodes are used for RBFN modeling. For SA-COSO, the sample sizes for its two populations are set to $d$ and $4d$, respectively. Note that for the experiments on 100-dimensional problems, we fix the number of training data size to $5d$ for FDD-EA by selecting the best samples for reducing the computation time.
	
	The experimental results are listed in Table \ref{Tab.S1}. As we can see, the proposed FDD-EA outperforms SA-COSO on 9 out of 10 benchmark problems according to the Wilcoxon test \cite{mann1947test} with the significance level being 0.05. It is undeperformed by SA-COSO only on the 100D Griewank function. 
	
\begin{figure*}[htb!]
		\centering
		\includegraphics[width=0.7\textwidth]{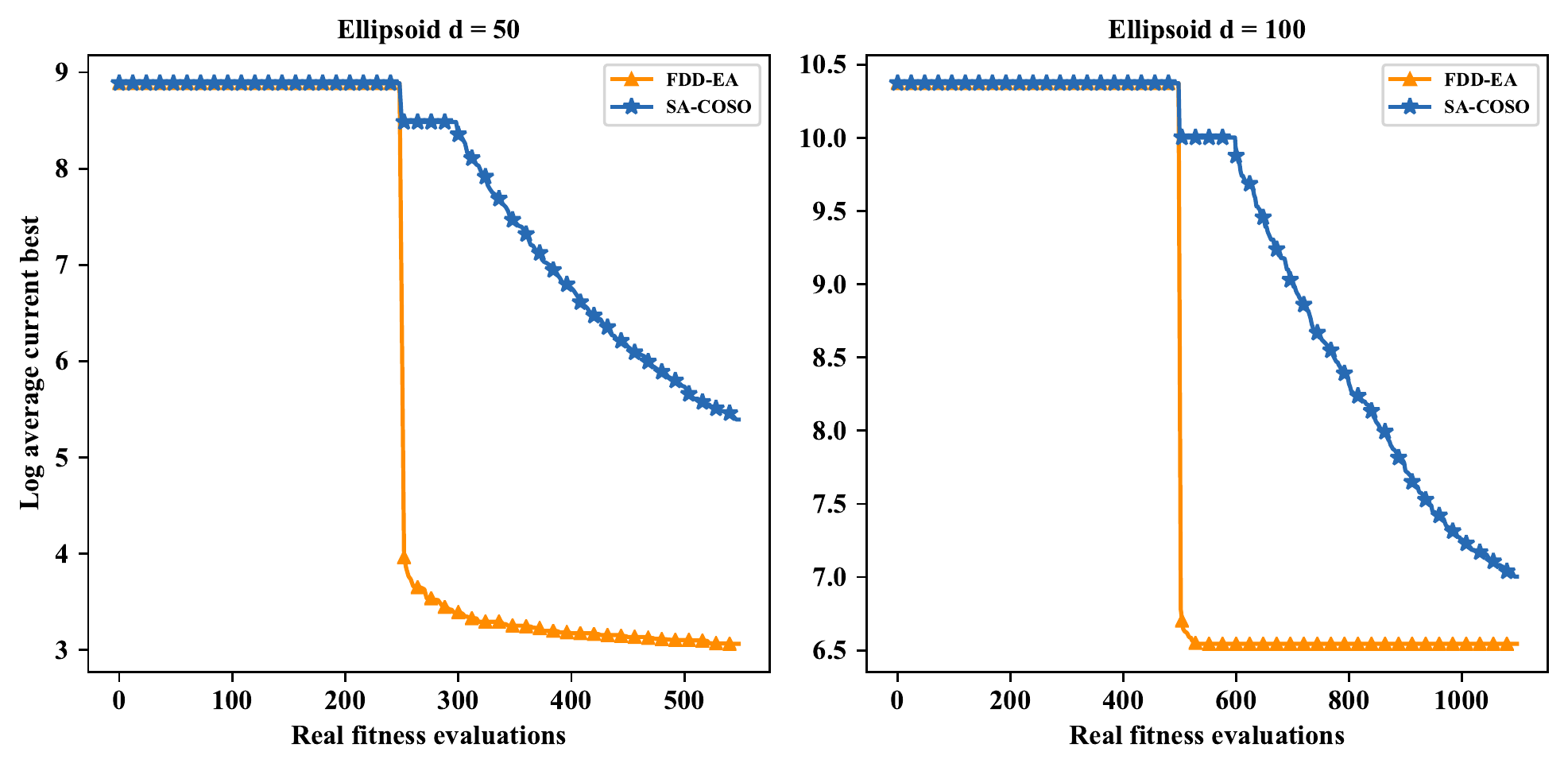}
		\caption{Convergence profiles of FDD-EA and SA-COSO on the Ellipsoid function in terms of the natural logarithm when $d=50, 100$.}
		\label{fig:E50_100}
	\end{figure*}

\begin{figure*}[htb!]
		\centering
		\includegraphics[width=0.7\textwidth]{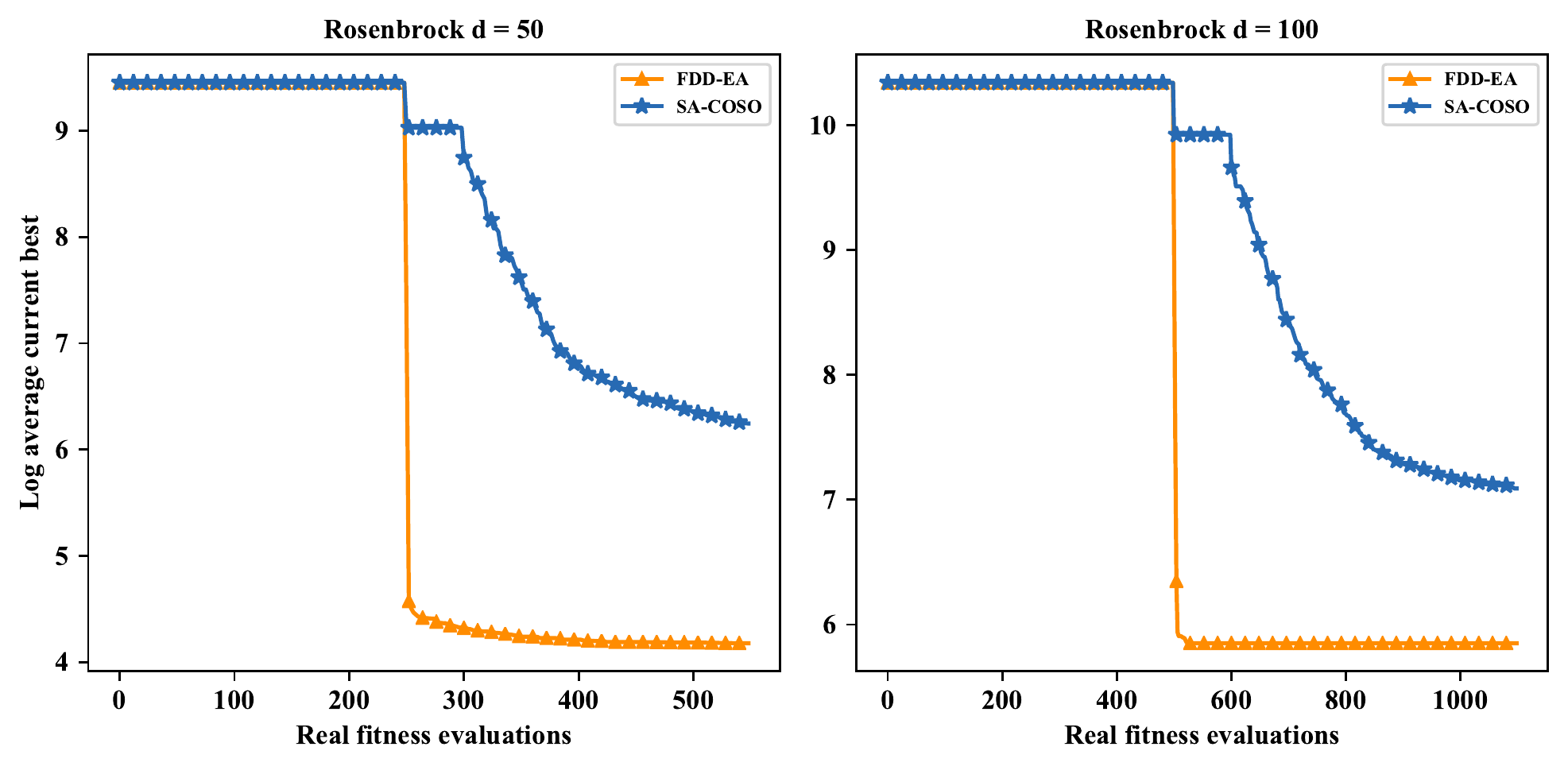}
		\caption{Convergence profiles of FDD-EA and SA-COSO in terms of the natural logarithm on the Rosenbrock function when $d=50, 100$.}
		\label{fig:Ro50_100}
	\end{figure*}

\begin{figure*}[htb!]
		\centering
		\includegraphics[width=0.7\textwidth]{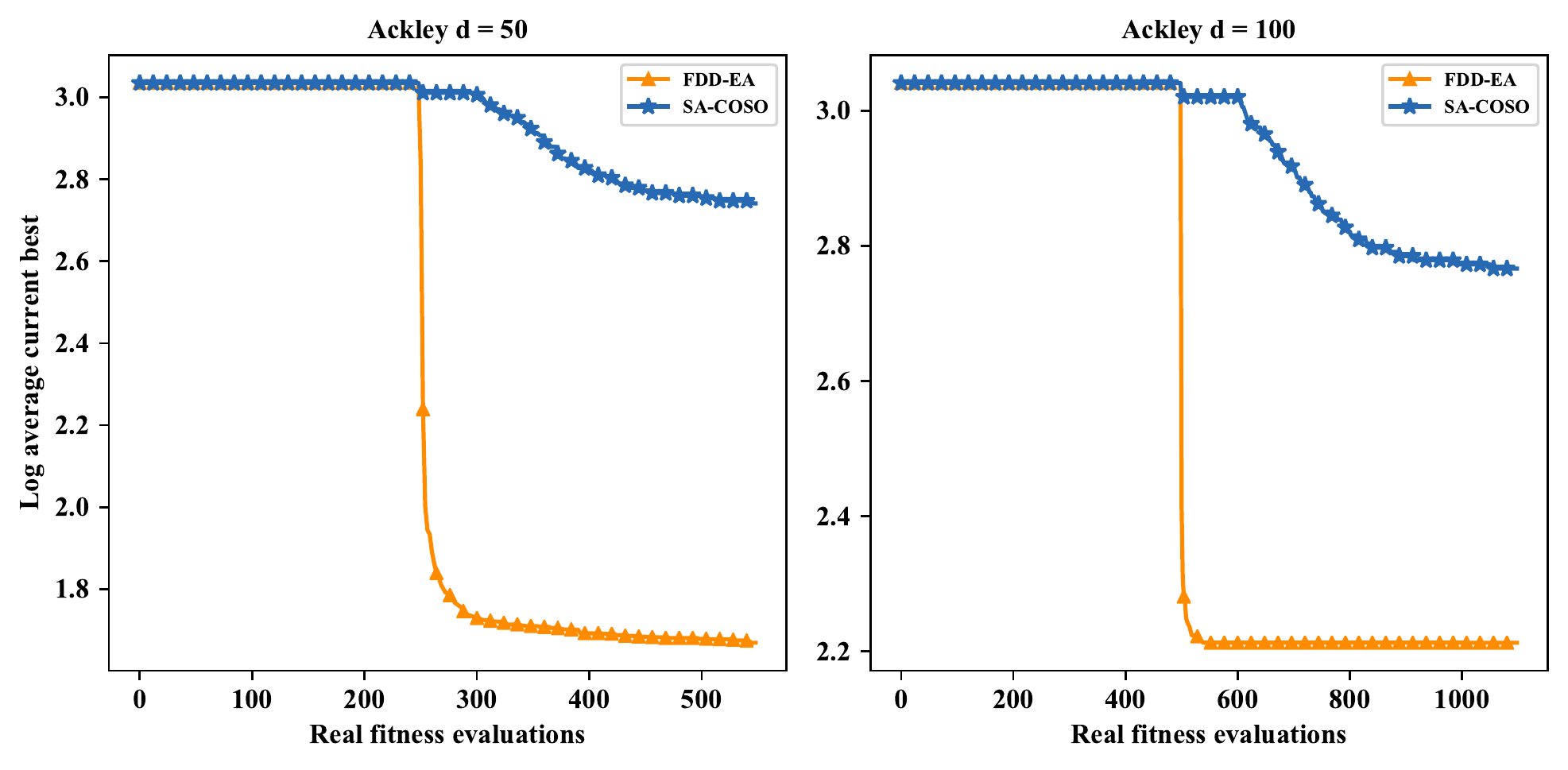}
		\caption{Convergence profiles of FDD-EA and SA-COSO in terms of the natural logarithm on the Ackley problem when $d=50, 100$.}
		\label{fig:A50_100}
	\end{figure*}

\begin{figure*}[htb!]
		\centering
		\includegraphics[width=0.7\textwidth]{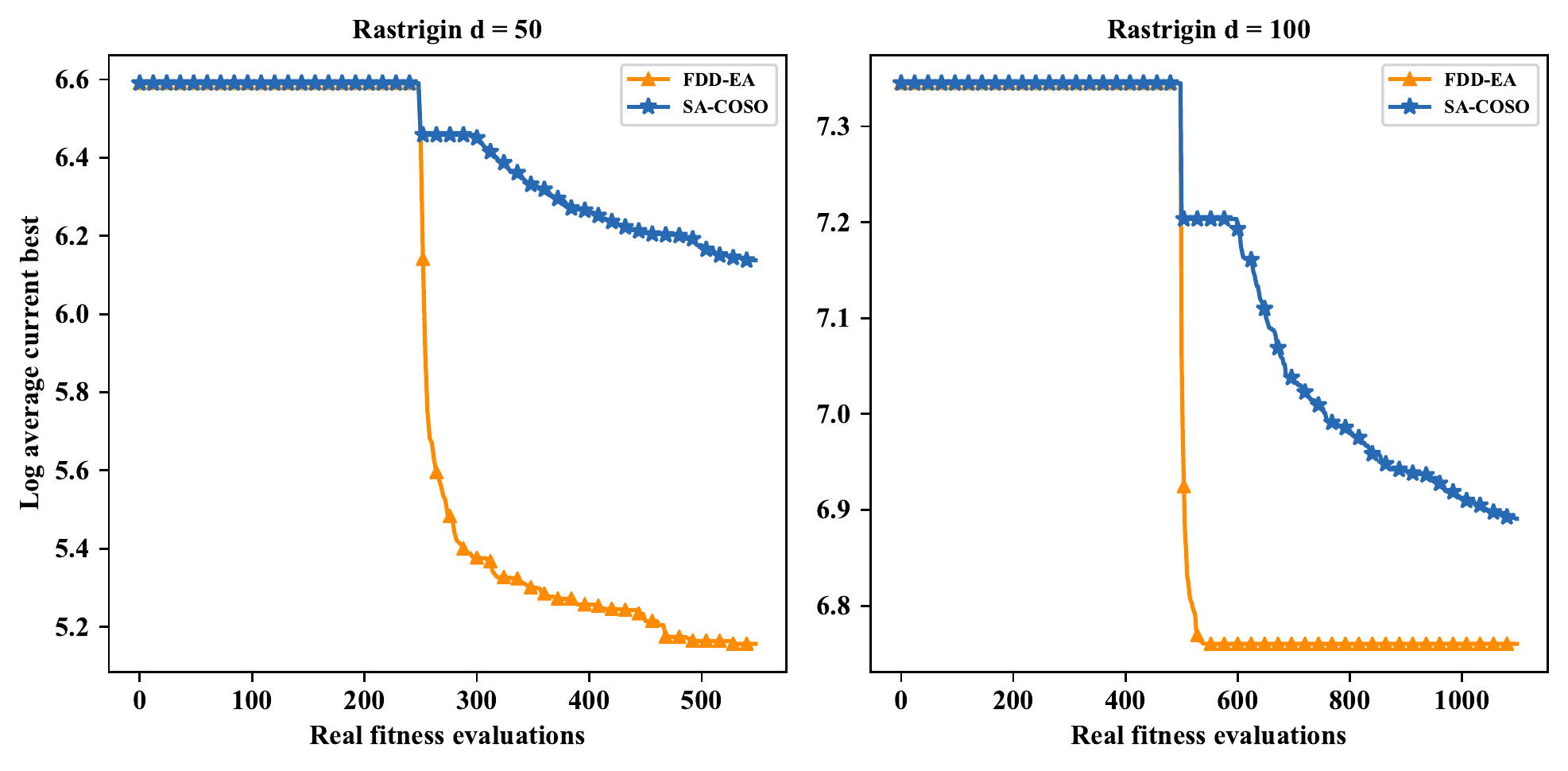}
		\caption{Convergence profiles of FDD-EA and SA-COSO in terms of the natural logarithm on the Rastrigin function when $d=50, 100$.}
		\label{fig:Ra50_100}
	\end{figure*}

\begin{figure*}[htb!]
		\centering
		\includegraphics[width=0.7\textwidth]{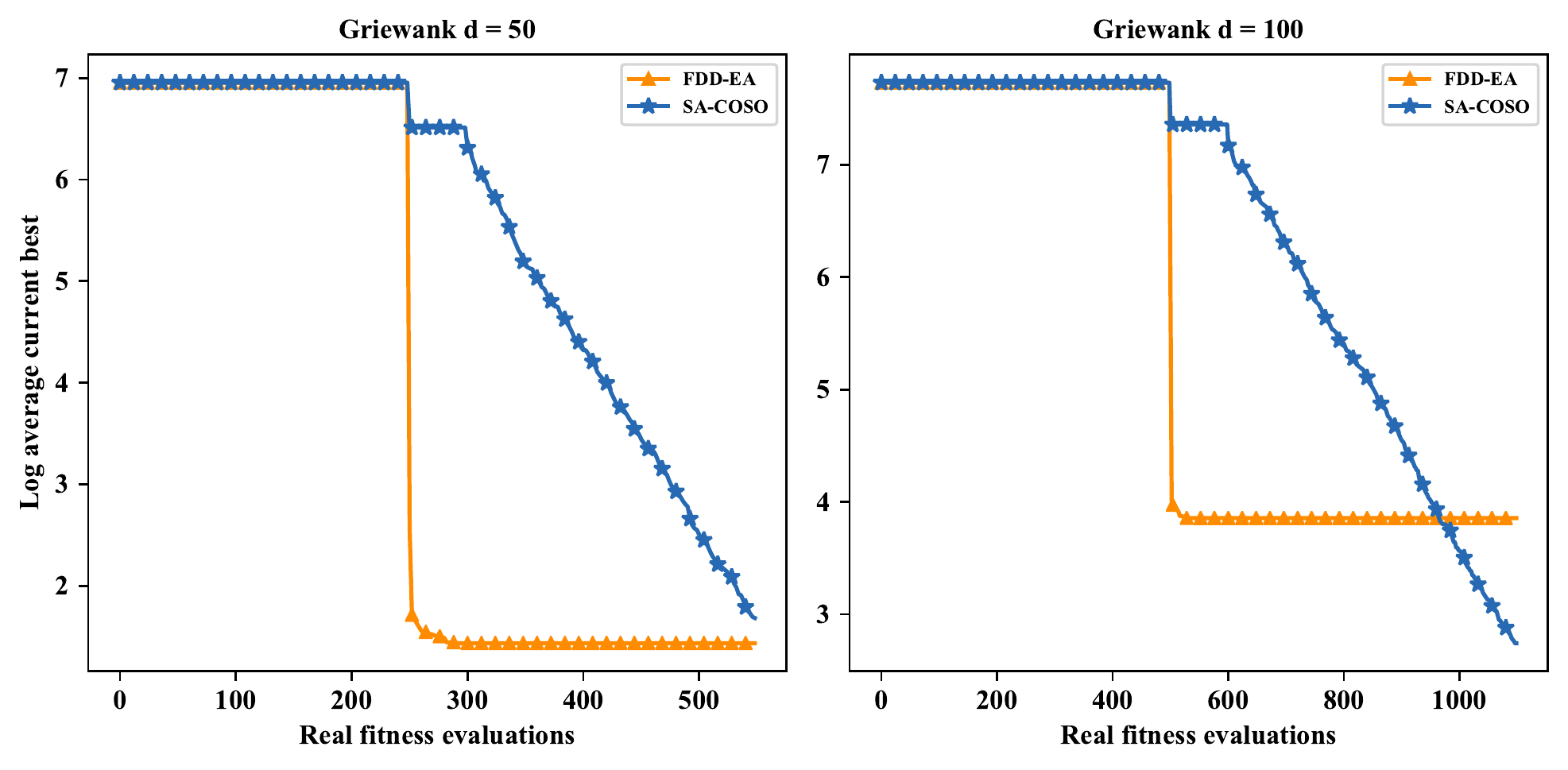}
		\caption{Convergence profiles of FDD-EA and SA-COSO in terms of the natural logarithm on the Griewank function when $d=50, 100$.}
		\label{fig:G50_100}
	\end{figure*}

	The convergence profiles of the two algorithms on the five benchmarks when $d=50,100$ are illustrated in Fig. \ref{fig:E50_100} - \ref{fig:G50_100}. Similar to lower dimensional cases, FDD-EA converges quickly on all test instances, and outperforms SA-COSO on four out of five test functions.  However, as shown in Fig. \ref{fig:G50_100}, the mean best fitness of SA-COSO continues to improve on the 50D and 100D Griewank function and eventually outperforms FDD-EA when the number of FE increases. This might be attributed to the fact that FDD-EA does not contain any fine search strategies as SA-COSO or other online SAEAs. 
	
	\begin{table*}[htb!]
		\caption{Average best fitness values (shown as avg $\pm$ std) on the benchmark functions when $d=50, 100$, where the $p$-values are calculated by the pairwise Wilcoxon rank sum test at 0.05 significance level.}
		
		\label{Tab.S1}
		\centering
		\renewcommand\arraystretch{1.5}
		\begin{tabular*}{0.85\textwidth}{@{\extracolsep{\fill}}ccccc}
			
			\hline
			Problem    & $d$ & FDD-EA                & SA-COSO        & $p$-value     \\ \hline
			\multirow{2}{*}{Ellipsoid}  
			& 50  & \textbf{2.14e+01 $\pm$ 2.43e+00} & 2.20e+02 $\pm$ 5.92e+01 &6.77e-08 (\textbf{+})
			\\
			& 100 & \textbf{6.95e+02 $\pm$ 5.40e+01} & 1.10e+03 $\pm$ 1.89e+02  &6.80e-08 (\textbf{+})
			\\ \hline
			
			\multirow{2}{*}{Rosenbrock}
			& 50  & \textbf{6.53e+01 $\pm$ 1.69e+00} & 5.16e+02 $\pm$ 7.78e+01 &6.53e-08 (\textbf{+})
			\\
			& 100 & \textbf{3.47e+02 $\pm$ 1.91e+01} & 1.20e+03 $\pm$ 1.49e+02 &6.79e-08 (\textbf{+})
			\\ \hline
			
			\multirow{2}{*}{Ackley}
			& 50  & \textbf{5.31e+00 $\pm$ 1.91e-01} & 1.55e+01 $\pm$ 5.93e-01 &6.12e-08 (\textbf{+})
			\\
			& 100 & \textbf{9.14e+00 $\pm$ 1.69e-01} & 1.59e+01 $\pm$ 4.01e-01  &6.73e-08 (\textbf{+})
			\\ \hline
			
			\multirow{2}{*}{Rastrigin}
			& 50  & \textbf{1.74e+02 $\pm$ 1.79e+01} & 4.63e+02 $\pm$ 3.56e+01 &6.68e-8 (\textbf{+})
			\\
			& 100 & \textbf{8.63e+02 $\pm$ 3.27e+01} & 9.83e+02 $\pm$ 4.88e+01  &6.79e-08 (\textbf{-})
			\\ \hline
			
			\multirow{2}{*}{Griewank}
			& 50  & \textbf{4.20e+00 $\pm$ 4.05e-01} & 5.38e+00 $\pm$ 1.76e+00 &2.00e-02 (\textbf{+})
			\\
			& 100 & 4.72e+01 $\pm$ 3.28e+00 & \textbf{1.55e+01 $\pm$ 3.65e+00}  &6.79e-08 (\textbf{-})
			\\ \hline
			
			Win / Lose / Tie & & 9/1/0 & 1/9/0
			\\ \hline
		\end{tabular*}
	\end{table*}
	
	%
	%

	\ifCLASSOPTIONcaptionsoff
	\newpage
	\fi

	
	
	%
	\bibliographystyle{IEEEtran}
	\bibliography{reference}

\end{document}